\documentclass[pdflatex,sn-mathphys-num]{sn-jnl}

\usepackage{graphicx}%
\usepackage{multirow}%
\usepackage{amsmath,amssymb,amsfonts}%
\usepackage{amsthm}%
\usepackage{mathrsfs}%
\usepackage[title]{appendix}%
\usepackage{xcolor}%
\usepackage{textcomp}%
\usepackage{manyfoot}%
\usepackage{booktabs}%
\usepackage{algorithm}%
\usepackage{algorithmicx}%
\usepackage{algpseudocode}%
\usepackage{listings}%
\usepackage{subcaption}
\newenvironment{psmallmatrix}
  {\left(\begin{smallmatrix}}
  {\end{smallmatrix}\right)}

\algrenewcommand\algorithmicrequire{\textbf{Input:}}
\algrenewcommand\algorithmicensure{\textbf{Output:}}

\theoremstyle{thmstyleone}%
\theoremstyle{thmstyletwo}%

\newcommand{\R}{\mathbb{R}}

\theoremstyle{thmstylethree}%

\begin{document}

\title[Weighted Low-Rank Matrix Approximation:\\ Acceleration and Applications]{Weighted Low-Rank Matrix Approximation:\\ Acceleration and Applications}


\author*[1]{\fnm{Elena} \sur{Tuzhilina}}\email{elena.tuzhilina@utoronto.com}

\author[2]{\fnm{Trevor} \sur{Hastie}}\email{hastie@stanford.edu}

\affil*[1]{\orgdiv{Department of Statistical Sciences}, \orgname{University of Toronto}}

\affil[2]{\orgdiv{Department of Statistics}, \orgname{Stanford University}}


\abstract{Weighted low-rank matrix approximation (WLRMA) generalizes classical low-rank approximation and matrix completion by allowing arbitrary elementwise weights. Such formulations arise naturally in a broad class of statistical models, including generalized linear low-rank models, where WLRMA serves as the computational primitive for parameter estimation. Despite its broad applicability, efficient optimization methods for general WLRMA remain relatively underdeveloped. In this paper, we formulate both the rank-constrained and nuclear-norm WLRMA problems within a unified first-order optimization framework by showing that the corresponding iterative algorithms are projected and proximal gradient methods. Building on this perspective, we develop accelerated algorithms based on Nesterov momentum and Anderson acceleration, together with a regularized Anderson scheme that improves numerical stability for non-convex problems. We further propose scalable  implementations for large sparse data matrices and introduce a practical
effective-rank criterion that provides a meaningful correspondence between rank-constrained and nuclear-norm solutions. We further show that fitting generalized linear low-rank models can be reduced to a sequence of weighted low-rank matrix approximation problems, allowing the proposed algorithms to be used as computational building blocks for their estimation. Simulation studies demonstrate substantial computational gains achieved by the proposed accelerated algorithms. Applications to the MovieLens dataset further illustrate the proposed framework for matrix completion, heteroscedastic Gaussian low-rank modeling, and logistic low-rank modeling.}




\maketitle

\section{Introduction}
\label{intro}

Low-rank matrix approximation is a cornerstone of multivariate statistics. Given a data matrix
$M\in\R^{n\times p}$, its best rank-$k$ approximation under squared error is obtained in
closed form by truncating the singular value decomposition (SVD)
\citep{eckart1936,mardia1979,greenacre2022}. When only a subset
$\Omega$ of the entries of $M$ is observed, the problem becomes one of low-rank
matrix completion,
\begin{equation*}
\min_{X\in\R^{n\times p}}
\sum_{(i,j)\in\Omega}(M_{ij}-X_{ij})^2
\quad\text{subject to}\quad
\operatorname{rank}(X)\le k,
\end{equation*}
where the summation is taken over the observed entries only.
Motivated in part by applications such as recommender systems and the Netflix Prize \citep{feuerverger2012},
matrix completion has become a central problem in statistics, machine learning, and
numerical optimization
\citep{candes2008,candes2010,feuerverger2012,nguyen2019}.

The matrix completion objective is a special case of a weighted objective $\sum_{i = 1}^n \sum_{j=1}^p W_{ij}(M_{ij} - X_{ij})^2$,
in which observed entries receive unit weight and missing entries receive zero weight. More generally,
one may assign arbitrary nonnegative weights $W_{ij}\in\mathbb{R}_+$ to individual matrix entries, leading to the weighted
low-rank matrix approximation (WLRMA) problem.

Weighted low-rank matrix approximation has a long history, dating at least to the work of \citet{gabriel1979}, who considered least-squares low-rank approximation under arbitrary elementwise weights. Subsequent work has investigated probabilistic formulations \citep{srebro2003}, the computational complexity of the problem \citep{gillis2011}, and theoretical properties of alternating-minimization algorithms \citep{li2016}. Nevertheless, unlike ordinary low-rank approximation, WLRMA admits no closed-form solution for general elementwise weights and is NP-hard in general \citep{gillis2011}. Practical methods therefore rely on iterative optimization algorithms. While matrix completion has led to a rich collection of scalable algorithms and theoretical guarantees, considerably less attention has been devoted to optimization methods for general elementwise-weighted low-rank approximation.

The importance of WLRMA extends well beyond matrix completion. Ordinary low-rank approximation is the maximum-likelihood estimator for a Gaussian model with a low-rank mean. More generally, a variety of statistical models have been proposed for matrix-valued data with non-Gaussian entry distributions, including exponential-family principal component analysis and generalized low-rank models \citep{collins2001,udell2016}. These frameworks have been successfully applied to a wide range of problems, including species abundance modeling in ecology \citep{robin2019,kidzinski2020} and modeling of chromatin contact data in computational biology \citep{tuzhilina2020,tuzhilina2024}, among many others. The optimization of these models can often be reduced to solving a sequence of weighted low-rank matrix approximation problems, allowing WLRMA algorithms to serve as computational building blocks for parameter estimation. Consequently, advances in algorithms for WLRMA directly translate into more efficient estimation procedures for a broad class of statistical models.

The contributions of this paper are summarized as follows.

First, we derive projected-gradient and proximal-gradient formulations for the rank-constrained and nuclear-norm formulations of WLRMA. This reinterpretation unifies the two classical approaches within a common first-order optimization framework and provides a natural foundation for acceleration. Building on this connection, we develop accelerated algorithms based on Nesterov momentum and Anderson acceleration. Although Anderson acceleration can substantially improve convergence, its direct application to the non-convex rank-constrained iteration can be unstable. We therefore introduce a regularized stabilization scheme that controls the Anderson mixing coefficients and improves the numerical stability of the accelerated method.

Second, we extend the proposed optimization framework to high-dimensional settings by developing scalable implementations based on sparse-plus-low-rank representations together with an alternating least-squares approximation of the proximal step. These implementations avoid repeated full singular value decompositions, making the proposed algorithms practical for large-scale problems.

Third, we derive an effective-rank criterion from the alternating least-squares formulation of WLRMA. Based on the degrees of freedom of ridge regression, the proposed criterion places rank-constrained and nuclear-norm approximations on a common scale, providing a practical tool for comparing hard- and soft-thresholding methods and selecting corresponding tuning parameters.

Finally, we establish a connection between generalized low-rank models and WLRMA by showing that many GLRM fitting procedures can be expressed as a sequence of weighted low-rank matrix approximation problems. This identifies WLRMA as a computational building block for fitting a broad class of statistical models.

We evaluate the proposed methods through simulation studies and on the MovieLens data. The experiments compare unaccelerated, Nesterov-accelerated, and Anderson-accelerated algorithms under both rank-constrained and nuclear-norm formulations, demonstrating the benefits of the proposed acceleration techniques and the regularized stabilization strategy. We further illustrate the use of the effective-rank criterion and demonstrate how WLRMA serves as a computational building block for fitting heteroscedastic Gaussian and logistic low-rank models. An \texttt{R} package implementing the proposed methods accompanies the paper.

The remainder of the paper is organized as follows. Section~\ref{sec:background} reviews the optimization formulations of WLRMA together with the classical hard- and soft-thresholding algorithms. Section~\ref{sec:pg} establishes their interpretation as projected-gradient and proximal-gradient methods. Section~\ref{sec:acceleration} develops accelerated algorithms based on Nesterov and Anderson acceleration and Section~\ref{sec:stabilized} introduces a regularized stabilization strategy for Anderson acceleration. Section~\ref{sec:highdim} presents scalable implementations for high-dimensional problems, and Section~\ref{sec:er} introduces the proposed effective-rank criterion. Section~\ref{sec:glr} demonstrates how WLRMA serves as a computational building block for generalized low-rank models. The remaining sections evaluate the proposed methods through simulation studies and illustrate their performance on the MovieLens dataset.

\section{Background}
\label{sec:background}

Throughout the paper, let $M\in\mathbb{R}^{n\times p}$ denote the observed data matrix and let
$W\in\mathbb{R}_+^{n\times p}$ be a matrix of nonnegative weights. We use $*$ to denote the Hadamard (elementwise) product, $\|\cdot\|$ the Euclidean vector norm,  $\|\cdot\|_F$ the Frobenius norm, and $\|\cdot\|_*$ the nuclear norm.

\subsection{Matrix completion as WLRMA}

Suppose only a subset $\Omega\subseteq[n]\times[p]$ of the entries of $M$ is observed, where
$[n]$ denotes the index set $\{1,\ldots,n\}$ (and similarly for $[p]$). Matrix completion seeks a low-complexity approximation $X\in\R^{n\times p}$ that agrees with the observed entries while ignoring the missing ones. 

The complexity of the solution is commonly controlled in two ways. The first directly constrains its rank, leading to the \emph{hard-thresholding formulation} 
\begin{equation}
\min_{X\in\mathbb{R}^{n\times p}}
\|P_\Omega(M-X)\|_F^2
\quad
\text{subject to}
\quad
\operatorname{rank}(X)\le k.
\label{eq:lrmc}
\end{equation}
Here $P_\Omega$ denotes the projection onto the observed entries, defined by
\[
(P_\Omega(X))_{ij}=
\begin{cases}
X_{ij},&(i,j)\in\Omega,\\
0,&(i,j)\notin\Omega.
\end{cases}
\]
The second replaces the rank constraint by a nuclear-norm penalty, yielding a convex relaxation that promotes low-rank solutions and leads to the \emph{soft-thresholding formulation}
\begin{equation}
\min_{X\in\R^{n\times p}}
\frac 12 \|P_\Omega(M-X)\|_F^2
+
\lambda\|X\|_*.
\label{eq:mc-nuclear}
\end{equation}

Both formulations naturally extend to weighted low-rank matrix approximation. Since
\[
P_\Omega(X)=W*X \qquad\text{with}\qquad
W_{ij}=
\begin{cases}
1,&(i,j)\in\Omega,\\
0,&(i,j)\notin\Omega,
\end{cases}
\]
the matrix completion objective is simply a weighted residual sum of squares with binary weights. Replacing these binary weights by arbitrary nonnegative weights yields the hard-thresholding formulation of weighted low-rank matrix approximation (WLRMA)
\begin{equation}
\min_{X\in\mathbb{R}^{n\times p}}
\|\sqrt{W}*(M-X)\|_F^2
\quad
\text{subject to}
\quad
\operatorname{rank}(X)\le k,
\label{eq:wlrma}
\end{equation}
and its soft-thresholding counterpart
\begin{equation}
\min_{X\in\R^{n\times p}}
\frac 12\|\sqrt{W}*(M-X)\|_F^2
+
\lambda\|X\|_*.
\label{eq:wlrma-nuclear}
\end{equation}

The weighted formulation encompasses matrix completion as the special case of binary weights while allowing arbitrary observation-specific weights arising in heteroscedastic regression, weighted matrix factorization, and, more generally, generalized low-rank models. 

\subsection{Thresholding algorithms}
\label{sec:algorithms}

The rank-constrained formulation \eqref{eq:lrmc} is commonly solved by
repeatedly imputing the missing entries and computing the best rank-$k$
approximation. The resulting iteration takes the form
\[
X^{(t+1)}
=
\operatorname{SVD}_k
\!\left(
P_\Omega(M)
+
P_{\Omega^\perp}(X^{(t)})
\right),
\]
where $\Omega^\perp=([n]\times[p])\setminus\Omega$ denotes the complement of the observed index set, and
$\operatorname{SVD}_k(Y)$ denotes the best rank-$k$ approximation of $Y$, obtained by truncating the singular value decomposition of $Y$ to its $k$ largest singular values. This operation is commonly referred to as \emph{singular-value hard thresholding}
\citep{cai2010}.

The convex relaxation \eqref{eq:mc-nuclear} is typically solved by
\emph{singular-value soft thresholding}
\citep{mazumder2010}. Specifically, let
\[
Y
=
U
\operatorname{diag}(d_1,\ldots,d_r)
V^\top
\]
be the singular value decomposition of
$Y\in\R^{n\times p}$, where $r=\min(n,p)$.
The operator
\[
\operatorname{SVT}_\lambda(Y)
=
U
\operatorname{diag}
\!\left(
(d_1-\lambda)_+,\ldots,(d_r-\lambda)_+
\right)
V^\top,
\]
obtained by shrinking each singular value by $\lambda$ and truncating at zero, is known as the \emph{singular-value soft-thresholding operator}, where $(a)_+=\max(a,0)$. Matrix completion is then solved by the iteration
\[
X^{(t+1)}
=
\operatorname{SVT}_\lambda
\!\left(
P_\Omega(M)
+
P_{\Omega^\perp}(X^{(t)})
\right).
\]

The corresponding algorithms for weighted low-rank matrix approximation replace the imputed matrix
$P_\Omega(M)+P_{\Omega^\perp}(X^{(t)})$
by the weighted surrogate
${W* M+(1-W)* X^{(t)}.}$
Applying $\operatorname{SVD}_k$ yields the EM-algorithm of
\citet{srebro2003}, while applying
$\operatorname{SVT}_\lambda$
produces algorithms for solving the weighted nuclear-norm formulation.

Although the hard- and soft-thresholding iterations are algorithmically similar, their connection to classical optimization methods is not immediately apparent. The next section shows that the weighted iterations are simply projected and proximal gradient methods applied to the rank-constrained and nuclear-norm formulations, respectively.

\section{Optimization viewpoint}
\label{sec:pg}

We briefly review proximal and projected gradient methods, which will provide a unified interpretation of the algorithms considered in this paper.
Many optimization problems can be formulated as composite optimization
problems of the form
\begin{equation}
\min_X g(X)+h(X),
\label{eq:pgd}
\end{equation}
where $g: \R^d\to\R$ is differentiable with Lipschitz-continuous gradient and $h: \R^d\to\R$ is
proper, closed, and convex
\citep{combettes2005,combettes2011,parikh2014}.
A standard approach for solving such problems is proximal gradient descent,
which generates the iterates
\[
X^{(t+1)}
=
\operatorname{prox}_{\eta h}
\!\left(
X^{(t)}-\eta\nabla g(X^{(t)})
\right),
\]
where $\eta>0$ is the step size and
\[
\operatorname{prox}_{\eta h}(Y)
=
\arg\min_X
\left\{
\frac12\|X-Y\|^2
+
\eta h(X)
\right\}.
\]
When $h$ is the indicator function of a closed convex set $C$,
\[
{1}_C(X)
=
\begin{cases}
0, & X\in C,\\
+\infty, & X\notin C,
\end{cases}
\]
the proximal operator reduces to the projection onto $C$,
\[
P_C(Y)
=
\operatorname{prox}_{\eta{1}_C}(Y)
=
\arg\min_{X\in C}\|X-Y\|,
\]
and the resulting algorithm is known as projected gradient descent.

\subsection{WLRMA optimization framework}

We now show that the thresholding algorithms of Section~\ref{sec:algorithms} arise naturally from the optimization framework introduced above. 
Throughout this section, let
\[
g(X)
=
\frac12
\|
\sqrt{W}*(M-X)
\|_F^2
\]
denote the weighted least-squares loss. Its gradient is given by
\[
\nabla g(X)
=
W*(X-M).
\]
Throughout the paper, we assume that the weights are normalized so that
$0 \le W_{ij} \le 1$ and $\max_{i,j}W_{ij}=1$. Consequently,
\[
\|\nabla g(X)-\nabla g(Y)\|_F
=
\|W*(X-Y)\|_F
\le
\|X-Y\|_F,
\]
so the gradient is Lipschitz continuous with constant
\(L=1\).

The rank-constrained and nuclear-norm WLRMA formulations correspond to two
different choices of $h$. As we show below, the resulting algorithms
perform the same gradient step and differ only in the operator applied
afterward. 

\subsection{Rank-constrained formulation}

Let
$C=\{X\in\R^{n\times p}:\operatorname{rank}(X)\le k\}$
denote the set of matrices of rank at most~$k$. Choosing
$h(X)={1}_C(X)$ in problem~\eqref{eq:pgd}
recovers the rank-constrained WLRMA problem~\eqref{eq:wlrma}. The gradient step is
\[
X^{(t)}-\eta\nabla g(X^{(t)})
=
(1-\eta W)*X^{(t)}+\eta W*M.
\]
Although $C$ is non-convex, the Eckart--Young theorem implies that a projection onto $C$ is available in closed form
$P_C(Y)=\operatorname{SVD}_k(Y).$
Hence projected gradient descent becomes
\[
X^{(t+1)}
=
\operatorname{SVD}_k
\!\left(
(1-\eta W)*X^{(t)}+\eta W*M
\right).
\]
For the unit step size $\eta=1$, this reduces to
\[
X^{(t+1)}
=
\operatorname{SVD}_k
\!\left(
W*M+(1-W)*X^{(t)}
\right),
\]
which is precisely the algorithm of \citet{srebro2003}. Thus, the classical weighted rank-constraint WLRMA algorithm can be viewed as projected gradient descent.

\subsection{Nuclear-norm formulation}

Choosing
$h(X)=\lambda\|X\|_*$
in problem~\eqref{eq:pgd} recovers the nuclear-norm
formulation~\eqref{eq:wlrma-nuclear}. The proximal operator of the
nuclear norm is the singular-value soft-thresholding operator
\citep{combettes2005,parikh2014}.
Therefore, the proximal-gradient iteration becomes
\[
X^{(t+1)}
=
\operatorname{SVT}_{\eta\lambda}
\!\left(
(1-\eta W)*X^{(t)}+\eta W*M
\right).
\]
For the unit step size $\eta=1$, this reduces to
\[
X^{(t+1)}
=
\operatorname{SVT}_{\lambda}
\!\left(
W*M+(1-W)*X^{(t)}
\right),
\]
which is the weighted soft-thresholding algorithm.

Since $\nabla g$ is Lipschitz continuous with constant
$L=1$, standard convergence results for proximal gradient descent
\citep{parikh2014} imply that, for any step size $0<\eta\le1$, the iterates
generated by the weighted soft-thresholding algorithm satisfy
\[
\ell(X^{(t)})-\ell(X^*)
\le
\frac{\|X^{(0)}-X^*\|_F^2}{2\eta t},
\]
where
\begin{equation}
\ell(X)
=
\frac12\|\sqrt{W}*(M-X)\|_F^2
+
\lambda\|X\|_*,    
\label{eq:loss}
\end{equation}
and $X^*$ is a minimizer of $\ell$. Thus, the objective values converge to
the optimum at the rate $\mathcal{O}(1/t)$. We note that no analogous convergence result is
guaranteed for the weighted hard-thresholding algorithm, since the
feasible set $C$ is non-convex. Throughout the remainder of
the paper, we set the gradient step to $\eta=1$.

\section{Acceleration}
\label{sec:acceleration}

The optimization interpretation developed in the previous section
immediately enables the use of classical acceleration techniques for
first-order methods. In this section, we consider two such approaches.
The first is Nesterov acceleration, which applies naturally to the
proximal-gradient formulation. The second is Anderson acceleration,
which can be applied to the corresponding fixed-point iterations. To
keep the presentation concise, we focus on the nuclear-norm
formulation. The accelerated algorithms for the rank-constrained
formulation are obtained analogously by replacing the singular-value
soft-thresholding operator $\operatorname{SVT}_{\lambda}$ with the
hard-thresholding operator $\operatorname{SVD}_k$.

\subsection{Nesterov acceleration}

Nesterov's accelerated gradient method is a widely used technique for
accelerating the convergence of first-order optimization algorithms
\citep{nesterov1983,beck2009}. The
method first constructs the extrapolated point
\[
V^{(t)}
=
X^{(t)}
+
\frac{t-1}{t+2}
(X^{(t)}-X^{(t-1)}),
\]
and then performs the proximal-gradient step from $V^{(t)}$, yielding the
iteration
\[
X^{(t+1)}
=
\operatorname{SVT}_{\lambda}
\!\left(
W*M+(1-W)*V^{(t)}
\right).
\]
This update is precisely a Nesterov-accelerated proximal-gradient
iteration applied to the weighted nuclear-norm formulation.
Consequently, the objective values satisfy
\[
\ell(X^{(t)})-\ell(X^*)
=
\mathcal O\!\left(\frac1{t^2}\right),
\]
improving upon the $\mathcal O(1/t)$ convergence rate of the standard
proximal-gradient method \citep{beck2009}. We note that this guarantee is specific to the convex nuclear-norm
formulation and does not automatically extend to the rank-constrained
formulation, since the corresponding optimization problem is nonconvex.

\subsection{Anderson acceleration}
\label{sec:anderson}

Anderson acceleration is a widely used technique for accelerating
fixed-point iterations \citep{anderson1965,walker2011}. Unlike
Nesterov acceleration, which constructs an extrapolated point using only
the two most recent iterates, Anderson acceleration adaptively combines
several previous updates by minimizing their fixed-point residuals.

To apply Anderson acceleration to the weighted soft-thresholding
algorithm, we first rewrite the proximal-gradient iteration as a
fixed-point equation by introducing the auxiliary variable
\begin{equation}
\label{eq:auxvar}
Y^{(t+1)}
=
W*M+(1-W)*X^{(t)},    
\end{equation}
which denotes the matrix obtained after the gradient step and before
singular-value thresholding. The updated estimate is then given by
$X^{(t+1)}=\operatorname{SVT}_{\lambda}(Y^{(t+1)})$. In the special case
of binary weights, $Y^{(t+1)}$ coincides with the imputed matrix.
The weighted soft-thresholding algorithm can therefore be viewed as
solving the fixed-point equation
\[
Y=f(Y)
\qquad\text{with}\qquad
f(Y)=W*M+(1-W)*\operatorname{SVT}_{\lambda}(Y).
\]

For implementation and notational convenience, we vectorize all
matrix-valued quantities. Let $\operatorname{vec}(\cdot)$ denote the
operator that stacks the columns of a matrix into a vector, and let
$\operatorname{mat}(\cdot)$ denote its inverse. At iteration $t$, define
\[
y^{(t)}=\operatorname{vec}(Y^{(t)}),\qquad
f^{(t)}=\operatorname{vec}\!\left(f(Y^{(t)})\right),\qquad
r^{(t)}=f^{(t)}-y^{(t)},
\]
where $f^{(t)}$ is the vectorized fixed-point update and $r^{(t)}$ is the
corresponding residual.

Let
$m_t=\min\{m,t\},$
where $m$ denotes the Anderson depth, and define
$
{R^{(t)}
=
\left(
r^{(t-m_t)},\ldots,r^{(t)}
\right),}
$
whose columns are the most recent $m_t+1$ residuals.
The Anderson coefficients are obtained by solving
\begin{equation}
\label{eq:anderson:alpha}
\min_{\alpha\in\mathbb R^{m_t+1}}
\|R^{(t)}\alpha\|^2
\qquad
\text{subject to}
\qquad
\alpha^\top\mathbf1=1.
\end{equation}
Similarly, define
$
F^{(t)}
=
\left(
f^{(t-m_t)},\ldots,f^{(t)}
\right),
$
whose columns are the corresponding fixed-point updates.
The Anderson iterate is then given by
\[
y^{(t+1)}=F^{(t)}\alpha^{(t)}.
\]
Provided that $R^{(t)\top} R^{(t)}$ is invertible, the solution to~\eqref{eq:anderson:alpha} is
\[
\alpha^{(t)}
=
\frac{
(R^{(t)\top} R^{(t)})^{-1}\mathbf1
}{
\mathbf1^\top
(R^{(t)\top} R^{(t)})^{-1}\mathbf1
}.
\]
In practice, the coefficients are computed by solving
$
R^{(t)\top} R^{(t)}\theta=\mathbf1
$
and normalizing
$
\alpha^{(t)}
=
\frac{\theta}{\mathbf1^\top\theta},
$
thereby avoiding explicit matrix inversion.
The resulting Anderson-accelerated weighted soft-thresholding algorithm
is summarized in Algorithm~\ref{alg:anderson-soft}.

\begin{algorithm}
\caption{WLRMA-AA}
\label{alg:anderson-soft}
\begin{algorithmic}[1]
\Require Data matrix $M$, weight matrix $W$, regularization parameter $\lambda$, Anderson depth~$m$
\State \textbf{Initialize:} $X^{(0)}=0$ and $Y^{(0)}=0$
\For{$t=0,1,2,\ldots$ until convergence}
    \State $m_t=\min\{m,t\}$
    \State $y^{(t)}=\operatorname{vec}(Y^{(t)})$
    \State $f^{(t)}=\operatorname{vec}\!\left(W*M+(1-W)*X^{(t)}\right)$
    \State $r^{(t)}=f^{(t)}-y^{(t)}$
    \State
    Form $R^{(t)}=(r^{(t-m_t)},\ldots,r^{(t)})$
    \State Find $\alpha^{(t)}$ by solving \eqref{eq:anderson:alpha}
    \State \hspace{\algorithmicindent}(a) Solve
    $R^{(t)\top}R^{(t)}\theta=\mathbf1$
    \State \hspace{\algorithmicindent}(b) Normalize
    $\alpha^{(t)}=\frac{\theta}{\mathbf1^\top\theta}$
    \State 
    Form $F^{(t)}=(f^{(t-m_t)},\ldots,f^{(t)})$
    \State $y^{(t+1)}=F^{(t)}\alpha^{(t)}$
    \State $Y^{(t+1)}=\operatorname{mat}(y^{(t+1)})$
    \State $X^{(t+1)}=\operatorname{SVT}_{\lambda}(Y^{(t+1)})$
\EndFor
\Ensure Low-rank estimate $X$
\end{algorithmic}
\end{algorithm}

The Anderson update admits a useful interpretation in relation to
Nesterov acceleration. Since the Anderson coefficients satisfy
$\mathbf{1}^\top\alpha^{(t)}=1$, the update of the auxiliary variable can
be written as
\begin{align}
Y^{(t+1)}
=
\operatorname{mat}\!\left(F^{(t)}\alpha^{(t)}\right) =
W*M
+
(1-W)*
\sum_{j=0}^{m_t}
\alpha_j^{(t)}X^{(t-m_t+j)}.
\label{anderson:update}
\end{align}
Defining the affine combination
$
V^{(t)}
=
\sum_{j=0}^{m_t}
\alpha_j^{(t)}X^{(t-m_t+j)},
$
the update becomes
$
Y^{(t+1)}
=
W*M+(1-W)*V^{(t)}.
$
Thus, both Anderson and Nesterov acceleration perform the
proximal-gradient step from an extrapolated point $V^{(t)}$. The
difference lies in how this point is constructed: Nesterov acceleration
uses a predetermined combination of the two most recent iterates,
whereas Anderson acceleration adaptively combines multiple previous
iterates by minimizing their fixed-point residuals.

The convergence theory of Anderson acceleration depends on the particular
variant under consideration. \citet{toth2015} establish local
$r$-linear convergence for fixed-depth Anderson acceleration when the
fixed-point mapping is contractive and the Anderson coefficients remain
uniformly bounded. More recently, \citet{zhang2020} establish global
convergence for a stabilized type-I Anderson method applied to
nonexpansive fixed-point mappings. Their algorithm incorporates
safeguarding, Powell-type regularization, and restarting, and therefore
differs from the Anderson-mixing formulation considered here. Their
results do not directly provide a convergence guarantee for
Algorithm~\ref{alg:anderson-soft}, but illustrate how suitable
stabilization mechanisms can restore global convergence for a modified
Anderson scheme.

In our implementation, we also consider several practical modifications of
the basic Anderson iteration. First, acceleration may be delayed for a
prescribed number of iterations, allowing the underlying proximal-gradient
method to establish a residual history before Anderson mixing is activated.
Second, an optional guarded strategy rejects an accelerated iterate whenever
it increases the objective function, reverting instead to the corresponding
proximal-gradient update. Finally, computing the Anderson coefficients requires forming the residual Gram matrix
$R^{(t)\top} R^{(t)}$. Recomputing this matrix from scratch at every
iteration costs $\mathcal{O}(npm^2)$, in addition to the
$\mathcal{O}(m^3)$ cost of solving the resulting linear system. Since only
one residual vector changes at each iteration, we instead update the Gram
matrix incrementally, reducing the Gram-matrix update cost to
$\mathcal{O}(npm)$. Because the Anderson depth $m$ is small in our
applications, solving the resulting $m$-dimensional linear system incurs
negligible additional cost. 

\section{Simulation}
\label{simulation}

In this section, we demonstrate the benefits of the proposed acceleration techniques through a small simulation example. We generate the data as follows. First, we draw $A^*\in\R^{n\times r}$ and $B^*\in\R^{p\times r}$ from the standard normal distribution and compute
$
M = A^*B^{*\top} + E,
$
where $E\in\R^{n\times p}$ is a matrix of errors drawn from $\mathcal N(0,\sigma^2)$. We further generate a matrix of weights $W\in\R^{n\times p}$ using a uniform distribution $W_{ij}\stackrel{\mathrm{i.i.d.}}{\sim}\operatorname{Unif}(0,1)$. In our experiments, we set $n=1000$, $p=100$, $r=75$, and $\sigma=1$.

We solve problems~(\ref{eq:wlrma}) and~(\ref{eq:wlrma-nuclear}) using
three algorithms: the baseline WLRMA, its Nesterov-accelerated variant, and its Anderson-accelerated variant. For Anderson acceleration, we set the depth parameter to $m=3$. When solving the non-convex WLRMA problem, we consider the solution ranks $k=10,25,50,75$. 
To ensure compatibility of results, for each rank $k$, we choose $\lambda$ so that the nuclear-norm solution has approximately the same weighted residual sum of squares $\|\sqrt{W}*(M-X)\|^2_F$ as the corresponding rank-constrained solution (we discuss a more formal way to match $\lambda$ and $k$ in Section~\ref{sec:er}). This results in the grid $\lambda=150,100,30,5$ for the penalty parameter.
For each method, we track the relative change in the objective
$
\Delta^{(t)}
=
\left|
\frac{\ell(X^{(t+1)})-\ell(X^{(t)})}
{\ell(X^{(t)})}
\right|,
$
where $\ell$ denotes the objective function corresponding to the optimized loss function under consideration. The algorithm is terminated once $\Delta^{(t)}<\epsilon$, where we set $\epsilon=10^{-8}$.

\begin{figure}[h!]
\centering
\includegraphics[width=0.99\textwidth]{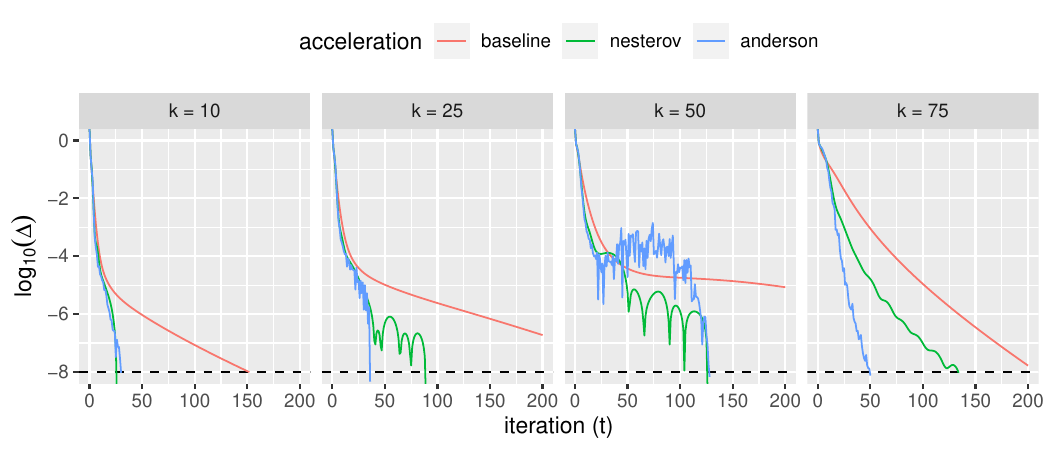}

\includegraphics[width=0.99\textwidth]{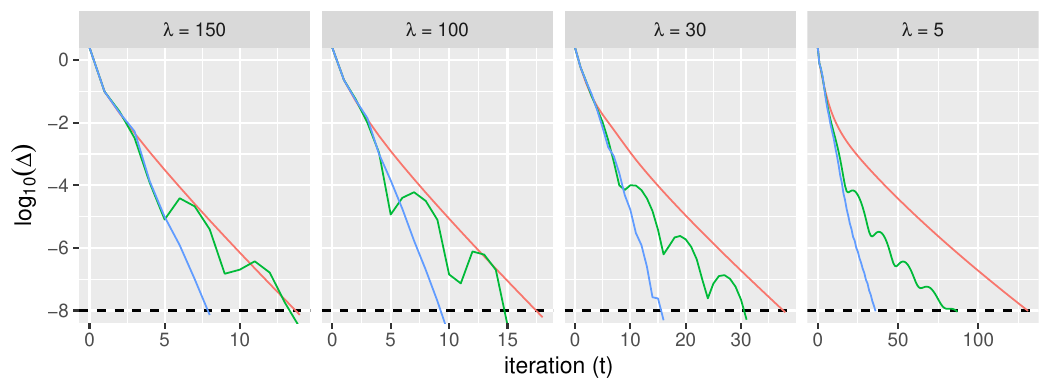}

\caption{Convergence of the baseline, Nesterov, and Anderson WLRMA algorithms. The curves show the relative objective change $\Delta^{(t)}$ versus iteration number for the rank-constrained (top) and convex (bottom) formulations. Anderson acceleration uses depth $m=3$.}
\label{fig:simulation:wlrma}
\end{figure}

According to Figure~\ref{fig:simulation:wlrma}, both acceleration methods converge substantially faster than the baseline algorithm. We also observe that, for the non-convex WLRMA problem, the trajectories of $\Delta^{(t)}$ for the accelerated methods are more erratic than those of the baseline, particularly for $k=50$. This behavior reflects the non-monotone sequence of iterates produced by the accelerated methods when applied to a non-convex optimization problem. Moreover, because of the non-convexity of the rank-constrained formulation, the choice of initialization may have a significant impact on the convergence behavior and the quality of the final solution. We investigate several initialization strategies in Appendix~\ref{app:simu:init}. 

\begin{figure}[h!]
\centering
\includegraphics[width=0.99\textwidth]{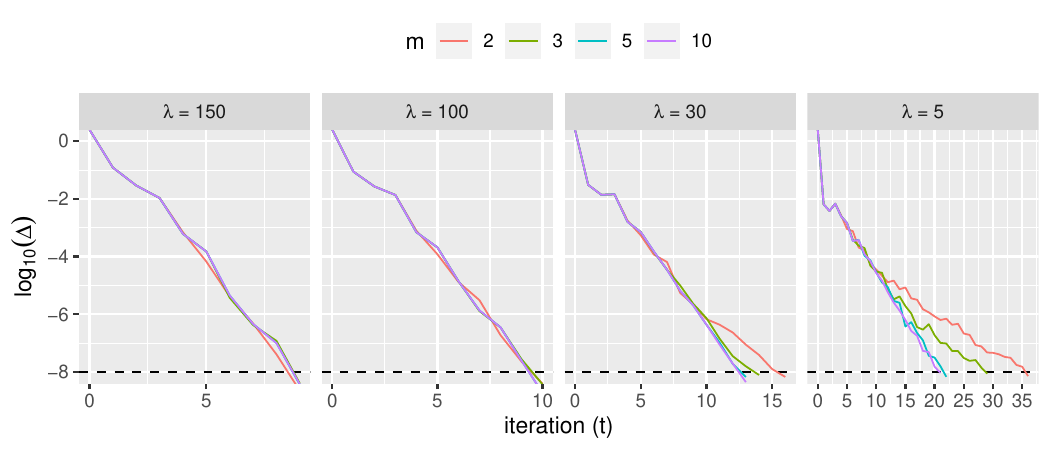}
\caption{Impact of the Anderson depth parameter on the convergence of the convex WLRMA algorithm. Larger values of $m$ provide only modest improvements, mainly for smaller values of $\lambda$.}
\label{fig:simulation:depth}
\end{figure}

To investigate the influence of the Anderson depth parameter, Figure~\ref{fig:simulation:depth} compares several values of $m$. We observe that increasing the depth provides only modest improvements in convergence, except for smaller values of $\lambda$, where larger depths lead to a more rapid decrease in $\Delta^{(t)}$ during the later iterations. Since these gains are relatively small compared with the additional computational costs, we use $m=3$ throughout the remainder of the paper.

Additional simulation results are presented in Appendix~\ref{app:simulation}. 
In particular, Figure~\ref{fig:simulation:init} demonstrates that, for the rank-constrained formulation,
a warm-start initialization substantially improves both the convergence speed and the stability
of the accelerated methods.
Figures~\ref{fig:simulation:rank} and~\ref{fig:simulation:noise} show that lower-rank matrices
and lower noise levels require fewer iterations for the convex WLRMA algorithm to converge.

\section{Stabilizing Anderson acceleration}
\label{sec:stabilized}

\begin{figure}[h!]
\centering

\includegraphics[width=\textwidth]{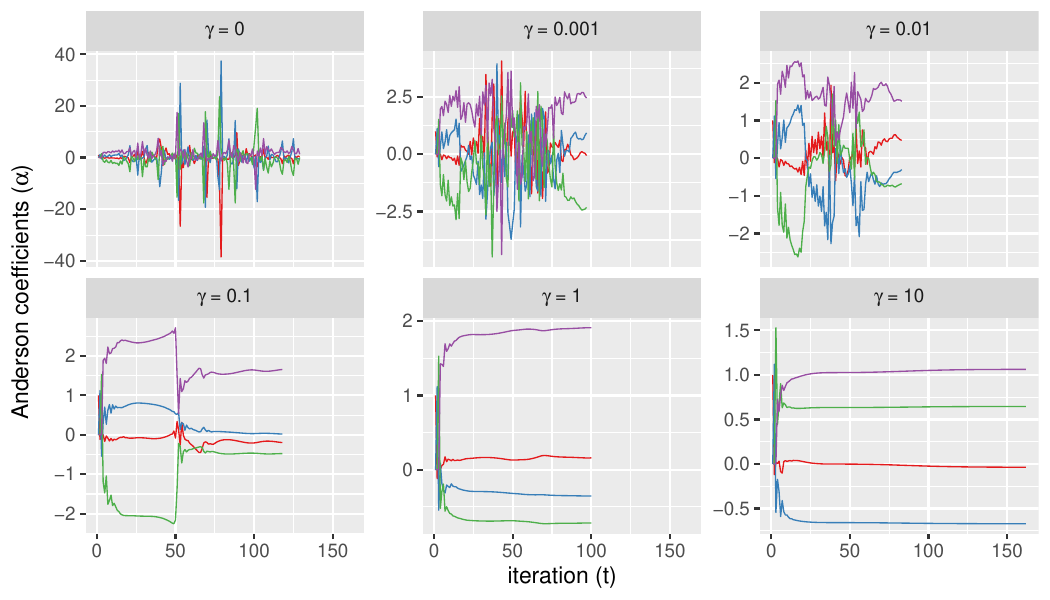}

\includegraphics[width=0.7\textwidth]{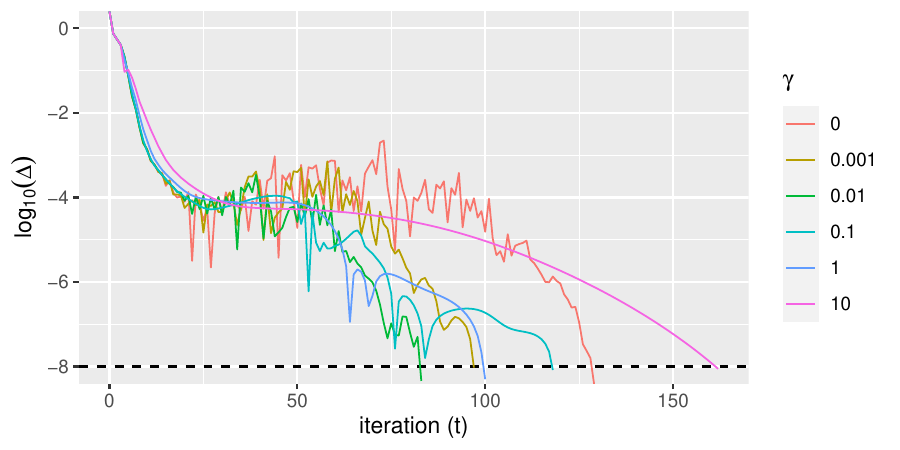}

\caption{
Stabilized Anderson acceleration for the rank-constrained WLRMA problem with solution rank $k=50$.
Top: Anderson coefficient trajectories for different values of the regularization parameter $\gamma$.
Bottom: Relative objective change $\Delta^{(t)}$ as a function of the iteration number.
}
\label{fig:simulation:compare}
\end{figure}

Although Anderson acceleration substantially improves convergence for the
nuclear-norm formulation of WLRMA, its application to the non-convex
rank-constrained problem may lead to erratic convergence behavior. This is
illustrated by the blue curve in the $k=50$ panel of
Figure~\ref{fig:simulation:wlrma}, where the objective change exhibits
pronounced oscillations. To better understand this phenomenon, we plot the
corresponding Anderson coefficient trajectories in the top panel of
Figure~\ref{fig:simulation:compare}. As shown in the $\gamma=0$ panel,
corresponding to the standard Anderson acceleration, the coefficients
themselves exhibit highly oscillatory behavior. To improve the
stability of Anderson acceleration, we propose a regularized coefficient
update.

Specifically, instead of solving problem~(\ref{eq:anderson:alpha}) we solve
\begin{align}
\label{eq:aa-reg}
\min_{\alpha\in\R^{m_t+1}}
\|R^{(t)}\alpha\|^2
+
\gamma^{(t)}
\|\alpha-\bar{\alpha}_d^{(t)}\|^2
\qquad
\text{subject to} 
\qquad
\alpha^\top\mathbf{1}=1
\end{align}
where
$\bar{\alpha}_d^{(t)}
=
\frac{1}{d}
\sum_{i=1}^{d}
\alpha^{(t-i)}$
is the average of the Anderson coefficient vectors from the previous
$d$ iterations, and $d$ is referred to as the regularization depth.
The additional penalty shrinks the new coefficient vector toward recent
values, thereby smoothing the coefficient trajectories while preserving
the affine constraint $\alpha^\top\mathbf1=1$. The solution to \eqref{eq:aa-reg} is
\begin{align*}
\alpha^{(t)}
=(I+\gamma^{(t)}K^{(t)})\,\alpha_0^{(t)}\quad\text{with}\quad
K^{(t)}
&=
\left(R^{(t)\top}R^{(t)}+\gamma^{(t)}I\right)^{-1}
\left(
\bar{\alpha}_d^{(t)}\mathbf{1}^\top
-
\mathbf{1}\bar{\alpha}_d^{(t)\top}
\right) \\
\text{and}\quad\alpha_0^{(t)}
&=
\frac{
\left(R^{(t)\top}R^{(t)}+\gamma^{(t)}I\right)^{-1}\mathbf{1}
}{
\mathbf{1}^\top
\left(R^{(t)\top}R^{(t)}+\gamma^{(t)}I\right)^{-1}
\mathbf{1}
}.
\end{align*}

Two practical implementation details deserve mention. We apply the regularized update only for $t\geq\max(m,d)$.
Before that point, we use the standard Anderson update.
Because the residual norms decrease during the iteration, we scale the
regularization parameter according to
$\gamma^{(t)}
=
\gamma
\|R^{(t)}\|_F^2,
$
where $\gamma>0$ is a user-specified constant. This scaling preserves a
comparable balance between the residual term and the regularization term
throughout the optimization.

The proposed regularization provides two practical benefits. First, it
suppresses large oscillations in the Anderson coefficients, leading to
more stable accelerated iterates. Second, as the fixed-point residuals
approach zero, the Gram matrix
$R^{(t)\top} R^{(t)}$ may become nearly singular. The regularization
improves its conditioning and therefore makes the coefficient computation
more numerically stable.

In our experiments, we set the Anderson depth to $m=3$, the regularization
depth to $d=3$, and considered the grid
$\gamma\in\{0,0.001,0.01,0.1,1,10\}$, where $\gamma=0$ recovers the standard
Anderson acceleration. Figure~\ref{fig:simulation:compare}
illustrates the effect of the proposed regularization on the challenging
rank-$50$ simulation example. Increasing $\gamma$ progressively smooths
the Anderson coefficient trajectories and reduces oscillations in the
convergence curves. Moderate values of $\gamma$ (approximately
$0.01$--$0.1$) provide the best trade-off between stability and
convergence speed, whereas excessively large values over-regularize the
update and slow convergence.

\section{High-dimensional implementation}
\label{sec:highdim}

When $n$ and/or $p$ are very large, repeatedly computing the truncated SVD
required by the projection step can become prohibitively expensive.
To address this issue, we adapt the alternating least squares (ALS)
strategy of \citet{hastie2015}, originally proposed for matrix
completion.

\subsection{Alternating least squares}
\label{sec:als}

The rank-constrained WLRMA problem~\eqref{eq:wlrma} can be equivalently written as
\begin{align}
\min_{A\in\R^{n\times k},\,B\in\R^{p\times k}}
\|\sqrt{W}*(M-AB^\top)\|_F^2,
\label{als:wlrma}
\end{align}
where $X=AB^\top$ is a rank-$k$ factorization of the solution. A standard
approach for solving \eqref{als:wlrma} alternates between optimizing $A$
and $B$, reducing each iteration to a collection of weighted least-squares
problems \citep{razenshteyn2016}. Although this avoids repeated
SVD computations, each alternating least squares (ALS) sweep requires solving $n+p$ independent
weighted least-squares problems, one for each row of $A$ and $B$.

Instead of applying ALS directly to \eqref{als:wlrma}, we apply it only to
the projection step of the projected gradient descent algorithm. Rather than computing the exact projection by truncated SVD, we approximate it using alternating least squares, resulting in an inexact projected-gradient iteration.
Recall
that, at iteration $t$, the projection is equivalent to finding the best
rank-$k$ approximation of
$
Y^{(t+1)}=W*M+(1-W)*X^{(t)},$
or, equivalently,
\[
\min_{A\in\R^{n\times k},\,B\in\R^{p\times k}}
\|Y^{(t+1)}-AB^\top\|_F^2.
\]
Unlike \eqref{als:wlrma}, this problem is unweighted. Consequently, each
ALS sweep consists of the closed-form updates
\begin{align}
B=Y^{(t+1)\top}A(A^\top A)^{-1}\qquad\text{and}\qquad
A=Y^{(t+1)}B(B^\top B)^{-1}.
\label{als:update}
\end{align}
This motivates the ALS-based approximation to the projected-gradient
iteration summarized in Algorithm~\ref{alg:als}. 
In practice, the target rank $k$ is typically chosen to be much smaller than both $n$ and $p$. Consequently, the factor matrices $A$ and $B$ are tall-and-skinny, and the updates in~\eqref{als:update} require matrix multiplications involving these matrices together with the solution of two (small) $k\times k$ linear systems.

\begin{algorithm}
\caption{WLRMA-ALS}
\label{alg:als}
\begin{algorithmic}[1]

\Require Data matrix $M$, weight matrix $W$, target rank $k$

\State \textbf{Initialize:} $A^{(0)}\in\mathbb{R}^{n\times k}$ and
$B^{(0)}\in\mathbb{R}^{p\times k}$ at random

\For{$t=0,1,2,\ldots$ until convergence}

\State 
$Y^{(t+1)}
=
W*M+(1-W)*A^{(t)}B^{(t)\top}$

\State 
$B^{(t+1)}
=
Y^{(t+1)\top}A^{(t)}
\bigl(A^{(t)\top}A^{(t)}\bigr)^{-1}$

\State Update
$Y^{(t+1)}
=
W*M+(1-W)*A^{(t)}B^{(t+1)\top}$

\State 
$A^{(t+1)}
=
Y^{(t+1)}B^{(t+1)}
\bigl(B^{(t+1)\top}B^{(t+1)}\bigr)^{-1}$

\EndFor

\Ensure Low-rank estimate
$X=AB^\top$

\end{algorithmic}
\end{algorithm}

Other variants of Algorithm~\ref{alg:als} are also possible. One may
omit Step~5 and construct $Y^{(t+1)}$ only once, using the same surrogate
for both factor updates, thereby reducing the computational cost of each
outer iteration. Alternatively, one may hold $Y^{(t+1)}$ fixed and
alternate between the updates of $A$ and $B$ (Steps~4 and~6) until
convergence, thereby recovering the standard ALS algorithm for computing
a rank-$k$ approximation of $Y^{(t+1)}$.

The same idea extends naturally to the convex formulation. Following
\citet{hastie2015}, for every $\lambda>0$ and sufficiently large $k$,
problem~\eqref{eq:wlrma-nuclear} is equivalent to 
\begin{align}
\min_{A\in\R^{n\times k},\,B\in\R^{p\times k}}
\frac12\|\sqrt{W}*(M-AB^\top)\|_F^2
+\frac{\lambda}{2}\|A\|_F^2
+\frac{\lambda}{2}\|B\|_F^2.
\label{als:cr:wlrma}
\end{align}
Thus, the proximal step
admits a factorized ridge-regression representation
\[
\min_{A\in\R^{n\times k},\,B\in\R^{p\times k}}
\frac 12\|Y^{(t+1)}-AB^\top\|_F^2+\frac{\lambda}{2}\|A\|_F^2
+\frac{\lambda}{2}\|B\|_F^2
\]
with updates
\begin{align}
B=Y^{(t+1)\top}A(A^\top A+\lambda I)^{-1}
\qquad\text{and}\qquad
A=Y^{(t+1)}B(B^\top B+\lambda I)^{-1}
\label{als:cr:update}
\end{align}
yielding an efficient alternating ridge regression algorithm for the nuclear-norm WLRMA formulation.

\subsection{Sparse weights}
\label{sec:sparse}

The ALS implementation developed in the previous section remains practical even
when the dimensions of the problem are very large. When the weight matrix is
also sparse, as in matrix completion case, the
computational and storage costs can be reduced even further. 

In this setting,
the surrogate matrix
$Y^{(t+1)}$
need not be formed explicitly.
Indeed,
\[
Y^{(t+1)}
=
\underbrace{W*(M-A^{(t)}B^{(t)\top})}_{S^{(t)}\ \text{(sparse)}}
+
\underbrace{A^{(t)}B^{(t)\top}}_{\text{low-rank}}.
\]
Thus, each surrogate matrix is represented as the sum of a sparse matrix and a
low-rank matrix. 
Using this representation, the ALS updates in
Algorithm~\ref{alg:als} can be implemented without explicitly forming
$Y^{(t+1)}$. Specifically, the update of $B$ is given by
\[
B^{(t+1)}
=
B^{(t)}
+
S^{(t)\top}
A^{(t)}
\left(A^{(t)\top}A^{(t)}\right)^{-1}.
\]
The sparse residual is then updated according to
$
S^{(t)}
=
W*\bigl(M-A^{(t)}B^{(t+1)\top}\bigr),
$
after which $A$ is updated as
\[
A^{(t+1)}
=
A^{(t)}
+
S^{(t)}B^{(t+1)}
\left(B^{(t+1)\top}B^{(t+1)}\right)^{-1}.
\]
For the nuclear-norm WLRMA formulation, the
ridge updates~\eqref{als:cr:update} can be evaluated using the same
sparse-plus-low-rank decomposition, without explicitly forming the
surrogate matrix. 

Consequently, both the rank-constrained and convex formulations can be
implemented using only multiplications involving sparse and tall-and-skinny
matrices together with the solution of two $k\times k$ linear systems per ALS
iteration. Moreover, the surrogate matrix is never formed explicitly.
Instead, only its sparse component $S^{(t)}$ on the support
$\Omega=\{(i,j):W_{ij}\neq0\}$ and the tall-and-skinny factor matrices
$A^{(t)}$ and $B^{(t)}$ are stored. As a result, the memory requirement
is reduced from $O(np)$ to $O((n+p)k+|\Omega|)$, allowing the algorithm
to scale to problems for which storing dense $n\times p$ matrices is
infeasible.

\subsection{Acceleration in high dimensions}
\label{sec:highdimacc}

The acceleration schemes developed in Section~\ref{sec:acceleration} extend
naturally to the high-dimensional ALS algorithms described in
Section~\ref{sec:highdim}. Let
$Z=\begin{psmallmatrix}A\\B\end{psmallmatrix}\in\mathbb{R}^{(n+p)\times k}$
denote the concatenated factor matrix, and let $\Phi$ denote one iteration of
Algorithm~\ref{alg:als} so that
$
Z^{(t+1)}=\Phi\!\left(Z^{(t)}\right).
$

Nesterov acceleration is obtained by replacing the current iterate with the
extrapolated point
\[
V^{(t)}
=
Z^{(t)}
+
\frac{t-1}{t+2}
\left(
Z^{(t)}-Z^{(t-1)}
\right)
\qquad\text{with}\qquad
Z^{(t+1)}
=
\Phi\!\left(V^{(t)}\right).
\]
Equivalently, the extrapolation is applied separately to the factor matrices
$A$ and $B$ before performing one ALS iteration.

Similarly, Anderson acceleration is obtained by applying the Anderson
acceleration scheme from Section~\ref{sec:anderson} to the fixed-point equation
$Z=\Phi(Z)$. Since the iterates consist only of the concatenated
factor matrix $Z\in\mathbb{R}^{(n+p)\times k}$ rather than the full
$n\times p$ surrogate matrix, the Anderson extrapolation and residual
computations are performed entirely in the low-dimensional factor space.
Consequently, the accelerated algorithm inherits the computational and
storage advantages of the underlying high-dimensional ALS implementation.

\section{Matching the rank-constrained and convex formulations}
\label{sec:er}

Both the rank-constrained and nuclear-norm WLRMA formulations require
selecting a tuning parameter, namely the target rank $k$ or the penalty
parameter $\lambda$. While the interpretation of $k$ is immediate, it
specifies the dimension of the latent subspace used to approximate the
data, the role of $\lambda$ is considerably less transparent. A natural
first attempt is to interpret the rank of the weighted
soft-thresholding solution as the corresponding latent dimension.
However, as we demonstrate below, this interpretation is misleading, as the algebraic rank
substantially overestimates the effective complexity of the fitted
model.

The discrepancy stems from the different thresholding operators employed
by the two formulations. The rank-constrained problem uses singular-value
hard thresholding, retaining only the largest $k$ singular values. In
contrast, the convex formulation applies singular-value soft
thresholding, replacing each singular value $d$ by $(d-\lambda)_+$.
Consequently, many singular values remain strictly positive while being
arbitrarily close to zero. Although these directions contribute
negligibly to the fitted model, they are still counted in the algebraic
rank, making it a poor measure of the intrinsic dimensionality of the
solution.

The alternating ridge regression formulation~\eqref{als:cr:wlrma}
provides a natural remedy to this problem. By viewing the nuclear-norm
formulation as an alternating ridge regression procedure, it suggests
measuring model complexity through the effective degrees of freedom of
the corresponding ridge regression problems. Specifically, recall that,
in ridge regression, the fitted responses can be written as
$\widehat y = Hy,$
where
${H=Z(Z^\top Z+\lambda I)^{-1}Z^\top}$
is the ridge hat matrix and $Z$ denotes the design matrix. The effective
degrees of freedom are then defined as
$edf(\lambda)=\operatorname{tr}(H),$
which measures the effective dimension of the fitted model
\citep{hastie2001}. Motivated by this analogy, we define an effective
rank for the nuclear-norm formulation.

Let $(A^*,B^*)$ denote a solution of the factorized problem
\eqref{als:cr:wlrma}. Fixing $A^*$, each row of $B^*$, viewed as a column
vector and denoted by $B_{j\cdot}\in\mathbb{R}^k$, is obtained by solving the
independent weighted ridge regression problem
\[
\min_{B_{j\cdot}\in\R^k}
\frac 12\|\sqrt{W_{\cdot j}}*(M_{\cdot j}-A^*B_{j\cdot})\|^2
+\frac\lambda2\|B_{j\cdot}\|^2.
\]
This suggests defining the effective rank of the nuclear-norm WLRMA solution as the
average effective degrees of freedom of these ridge regressions,
\begin{align}
er(\lambda)
=
\frac1p
\sum_{j=1}^p
\operatorname{tr}
\left[
(A^{*\top}D_jA^*+\lambda I)^{-1}
A^{*\top}D_jA^*
\right],
\label{er}
\end{align}
where $D_j=\operatorname{diag}(W_{\cdot j})$. An analogous definition can be
obtained by fixing $B^*$ instead of $A^*$. As demonstrated in
Appendix~\ref{app:er}, both definitions produce nearly identical effective-rank
estimates.

To illustrate the proposed definition, we return to the simulation setting of
Section~\ref{simulation}. We solve the nuclear-norm WLRMA problem over the grid
$\lambda=25,50,\ldots,200$ and record the corresponding weighted residual sums
of squares. For each value of $\lambda$, we additionally compute the algebraic
rank of the solution and the proposed effective rank
$er(\lambda)$. We then solve the rank-constrained WLRMA problem twice: first
using the algebraic rank of the convex solution as the target rank, and second
using the effective rank rounded to the nearest integer.

\begin{figure}[h!]
    \centering
    \includegraphics[width=.8\textwidth]{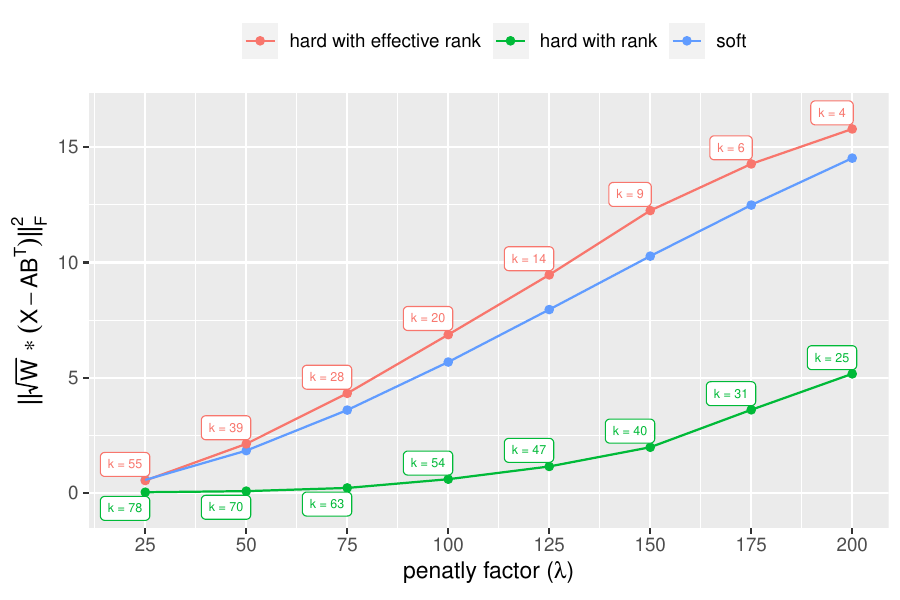}
    \caption{Comparison of the algebraic and proposed effective ranks for
matching the nuclear-norm and rank-constrained WLRMA formulations.
The blue curve corresponds to the nuclear-norm formulation, while the
green and red curves correspond to the rank-constrained formulation with
$k$ chosen as the algebraic and proposed effective ranks,
respectively. The selected values of $k$ are indicated in the figure.}
    \label{fig:df}
\end{figure}

Figure~\ref{fig:df} compares the resulting approximation errors and demonstrates that the algebraic rank substantially overestimates the
effective dimension of the nuclear-norm solution. When this rank is used in the
rank-constrained formulation, the resulting approximation errors are
considerably smaller than those of the soft-thresholding WLRMA estimator. In contrast, replacing
the algebraic rank by the proposed effective rank yields approximation errors
that closely match those of the nuclear-norm formulation. Although the effective rank
slightly underestimates the latent dimension in this example, it provides a much
more meaningful correspondence between the tuning parameters $\lambda$ and $k$
than the algebraic rank of the convex solution.

\section{MovieLens example}
\label{realdata}

We first illustrate the proposed accelerated WLRMA algorithms in the classical matrix completion setting, where the weight matrix is binary and indicates the observed entries. We use the MovieLens 1M dataset, which contains one million ratings from 6,000 users on 4,000 movies. The data are represented as a sparse rating matrix $M\in\mathbb{R}^{6000\times4000}$, in which only about 5\% of the entries are observed. The goal is to recover a low-rank approximation of $M$ from these partially observed ratings. In the following section, we consider the more general weighted setting with non-binary observation weights.

This dataset provides a realistic large-scale matrix completion problem and is therefore well suited for evaluating the computational performance of the proposed algorithms. We exploit the sparse structure of the binary weight matrix and use the high-dimensional ALS implementation from Section~\ref{sec:highdim}, which operates directly on the sparse observations without explicitly forming dense matrices, substantially reducing both computational and memory costs.

\begin{figure}[h]
  \centering
  \begin{subfigure}{\textwidth}
    \includegraphics[width=\textwidth]{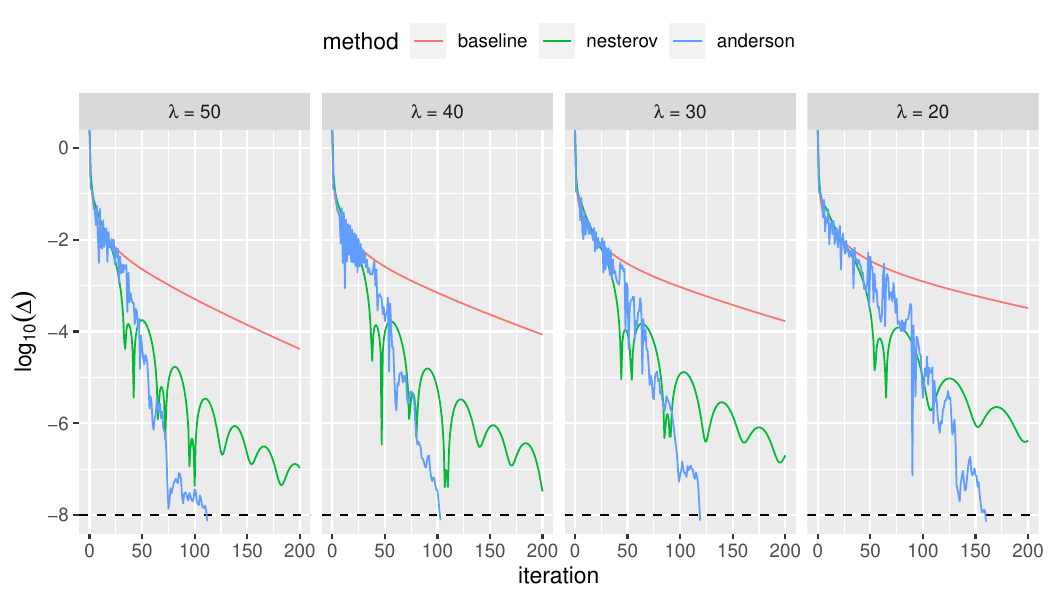}
  \end{subfigure}

  \begin{subfigure}{\textwidth}
    \includegraphics[width=\textwidth]{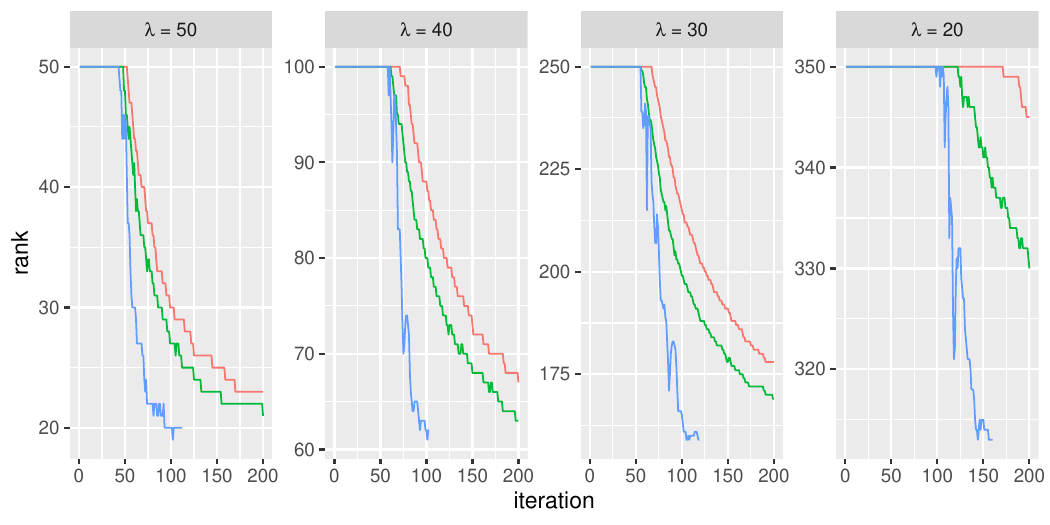}
  \end{subfigure}

  \caption{Comparison of the baseline ALS algorithm with Nesterov and Anderson acceleration on the MovieLens dataset. The convex WLRMA problem is solved for $\lambda=20,30,40,$ and $50$. Top: relative change in the objective function. Bottom: rank of the current iterate.}
  \label{fig:realdata:compare}
\end{figure}

We compare three optimization strategies: the baseline ALS algorithm, ALS with Nesterov acceleration, and ALS with Anderson acceleration using depth $m=3$. All methods are initialized using a warm start obtained from the unweighted low-rank matrix approximation of the rating matrix, where missing entries are replaced by zeros. For Anderson acceleration, we use the guarded strategy, accepting an accelerated iterate only when it decreases the objective. We consider four values of the regularization parameter, $\lambda=20,30,40,50$, corresponding to solutions of different complexity. All algorithms are terminated when the relative change in the objective function falls below $\epsilon=10^{-8}$ or after a maximum of 200 iterations.

Figure~\ref{fig:realdata:compare} compares the convergence of the baseline ALS algorithm with its Nesterov- and Anderson-accelerated variants for four values of the regularization parameter. The top panel reports the relative change in the objective function, while the bottom panel shows the algebraic rank of the current iterate.
Both acceleration techniques substantially reduce the iteration number required to reach a prescribed level of accuracy compared with the baseline ALS algorithm. Nesterov acceleration achieves a rapid initial decrease in the objective function, whereas Anderson acceleration converges more consistently and reaches the stopping criterion first for all four values of $\lambda$. These results demonstrate that both acceleration strategies provide considerable practical improvements for large-scale sparse matrix completion, with Anderson acceleration being particularly effective when high numerical accuracy is required.

The four values of the regularization parameter produce solutions with algebraic ranks $313$, $159$, $61$, and $19$, respectively. However, the corresponding effective ranks introduced in Section~\ref{sec:er} are substantially smaller, namely $29$, $12$, $5$, and $3$. This discrepancy indicates that the algebraic rank may considerably overestimate the intrinsic complexity of the fitted model and highlights the usefulness of the proposed effective-rank criterion for interpreting and selecting regularized low-rank solutions.

\section{Applications of WLRMA to statistical modeling}
\label{sec:glr}

The WLRMA framework extends naturally beyond matrix completion and can serve as a computational building block for fitting a broad class of statistical models. In many applications, the observed matrix is viewed as a realization of random variables whose distribution depends on an unknown low-rank parameter matrix. Estimation is then formulated as the maximization of a likelihood or, more generally, the minimization of a loss function under a low-rank constraint. As we show below, a variety of such optimization problems can be solved efficiently by repeatedly reducing them to weighted low-rank matrix approximation problems.
In this section, we illustrate this principle using two examples. The first extends the standard Gaussian model for observed entries by allowing heteroscedastic noise. The second considers a logistic low-rank model for the missing pattern.

\subsection{Heteroscedastic Gaussian model}
\label{sec:heteroscedasticity}

Ordinary matrix completion can be interpreted as maximum likelihood estimation under a homoscedastic Gaussian model for the observed entries. A natural extension is to allow the observation variance to vary across entries. Specifically, suppose that
$
M_{ij}\sim \mathcal N(X_{ij},\sigma_{ij}^2)$ for $(i,j)\in\Omega,$
where $X$ is an unknown low-rank mean matrix and $\sigma_{ij}^2$ are known variances. The negative log-likelihood is, up to an additive constant,
\[
\frac 12
\sum_{(i,j)\in\Omega}
\left(
\log\sigma_{ij}^2+
\frac{(M_{ij}-X_{ij})^2}{\sigma_{ij}^2}
\right).
\]
Consequently, estimating $X$ is equivalent to solving a WLRMA problem with weights
\begin{equation}
W_{ij}=
\begin{cases}
0, & (i,j)\notin\Omega,\\[0.3em]
\frac1{\sigma_{ij}^2}, & (i,j)\in\Omega.
\end{cases}
\label{eq:weights}    
\end{equation}

As an illustration, we apply this model to the MovieLens dataset and assume that the observation variance depends only on the user. Specifically, we let $\sigma_{ij}^2=\sigma_i^2$, so that each user has a common variance across all rated movies. This assumption reflects the idea that some users provide more consistent ratings than others while remaining simple enough to estimate reliably.

A simple approach is to estimate the user-specific variances directly from the observed ratings. Let
$p_i=\#\{j:(i,j)\in\Omega\}$
denote the number of ratings provided by user $i$. We use the within-user rating variance
\begin{equation}
\label{eq:s2}
\hat\sigma_i^2=
\frac{1}{p_i-1}
\sum_{j:(i,j)\in\Omega}
(M_{ij}-\bar M_i)^2
\qquad\text{with}\qquad
\bar M_i=\frac{1}{p_i}\sum_{j:(i,j)\in\Omega}M_{ij}.
\end{equation}
as an initial estimate of $\sigma_i^2$. These estimates are used to construct the weight matrix~\eqref{eq:weights}, which is subsequently used to estimate the low-rank mean matrix $X$ via WLRMA.

The approach described above estimates the variances directly from the observed ratings while ignoring the low-rank structure of the mean matrix. Consequently, the estimated variances reflect both the underlying signal and the observation noise. A more principled approach is therefore to estimate the mean matrix $X$ and the variances $\{\sigma_i^2\}$ jointly.

Given a low-rank estimate $X$, the maximum likelihood estimator of $\sigma_i^2$ is
\begin{equation}
\hat\sigma_i^2=
\frac{1}{p_i}
\sum_{j:(i,j)\in\Omega}
(M_{ij}-X_{ij})^2.
\label{sigma:adaptive}
\end{equation} 
This observation naturally leads to an alternating optimization procedure that repeatedly updates the low-rank mean matrix using WLRMA with the current weights and then recomputes the variances from the resulting residuals. The resulting algorithm is summarized in Algorithm~\ref{alg:adaptive}.

\begin{algorithm}
\caption{Adaptive WLRMA}
\label{alg:adaptive}
\begin{algorithmic}[1]

\Require data $M$, target rank $k$

\State \textbf{Initialize:} Compute $(\hat\sigma_i^{(0)})^2$ using
\eqref{eq:s2}, for $i=1,\ldots,n$

\For{$t=0,1,2,\ldots$ until convergence}

\State Set
$W_{ij}^{(t)}=\frac1{(\hat\sigma_i^{(t)})^2}$ for $(i,j)\in\Omega$
and $W_{ij}^{(t)}=0$ otherwise

\State Compute
$X^{(t+1)}$ by solving \eqref{eq:wlrma} with data $M$
and weights $W^{(t)}$

\State Update variance
$(\hat\sigma_i^{(t+1)})^2
=
\frac 1 {p_i}\sum_{j:(i,j)\in\Omega}
(M_{ij}-X_{ij}^{(t+1)})^2$,
for $i=1,\ldots,n$

\EndFor

\Ensure Mean estimate $X$ and variance estimates
$\{\hat\sigma_i^2\}_{i=1}^n$

\end{algorithmic}
\end{algorithm}

The proposed methods are evaluated on the MovieLens dataset. To reduce the sparsity of the data, we restrict the analysis to movies rated by at least 300 users and users who rated at least 200 movies. The resulting rating matrix has dimensions $1201\times1053$ and contains approximately $67\%$ missing entries.

To compare the different weighting strategies, we perform ten-fold cross-validation using the observed ratings. In each split, nine folds are used to estimate a rank-$k$ low-rank approximation, while the remaining fold is reserved for evaluation. We compare three models: the homoscedastic model corresponding to binary weights, the heteroscedastic model with user-specific variances estimated from \eqref{eq:s2}, and the adaptive heteroscedastic model obtained by Algorithm~\ref{alg:adaptive}. Prediction accuracy is measured by the residual sum of squares on the held-out observations. Since the true rank is unknown, we consider $k=2,4,\ldots,20$.

\begin{figure}[h]
    \centering

    \begin{subfigure}[b]{0.37\textwidth}
        \centering
        \includegraphics[width=\textwidth]{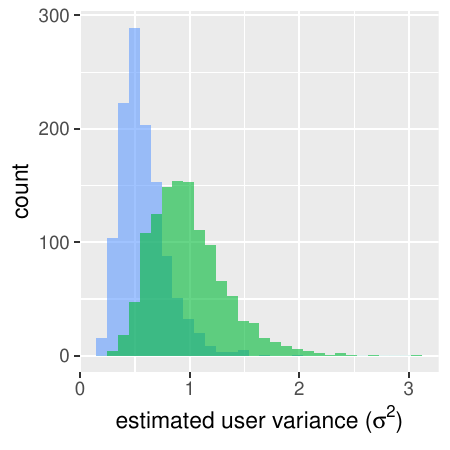}
    \end{subfigure}
    \hfill
    \begin{subfigure}[b]{0.60\textwidth}
        \centering
        \includegraphics[width=\textwidth]{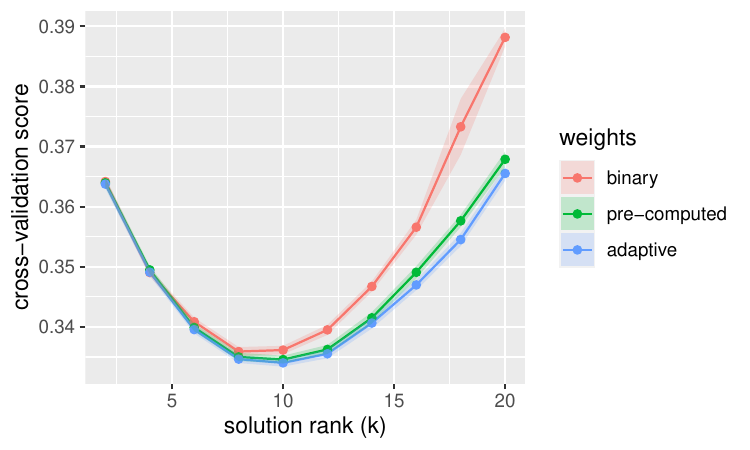}
    \end{subfigure}

    \caption{
Comparison of homoscedastic and heteroscedastic Gaussian models on the MovieLens dataset.
Left: estimated user-specific variances for the fixed and adaptive procedures.
Right: ten-fold cross-validation residual sum of squares versus approximation rank.
}
    \label{fig:hetero}
\end{figure}

Figure~\ref{fig:hetero} summarizes the results. The heteroscedastic models consistently outperform the homoscedastic model across all values of $k$, demonstrating the benefit of accounting for user-specific variability. Moreover, jointly estimating the low-rank mean matrix and the variances yields a further improvement over using the fixed variance estimates obtained from \eqref{eq:s2}. The adaptive procedure achieves the lowest prediction error at rank $k=10$.

The left panel of Figure~\ref{fig:hetero} compares the user-specific variance estimates obtained from the fixed and adaptive procedures for the final rank-$10$ solution. Compared with the fixed estimates, the adaptive estimates are shifted toward smaller values, indicating that the low-rank model explains part of the variability that is attributed to noise when the variances are estimated directly from the observed ratings.

It is worth noting that, in all the experiments, we set the convergence threshold for WLRMA to a relatively small value of $\epsilon = 10^{-6}$. Although the model involving adaptive weights required several runs of WLRMA, the implementation of WLRMA-ALS with sparse weights and acceleration allowed us to obtain the solution within just a few seconds.

\subsection{Generalized low-rank models}
\label{sec:glm}

Many applications involve observations that are not adequately modeled by Gaussian distributions. Examples include binary data arising from implicit feedback in recommendation systems, count data in genomics, and categorical responses in survey analysis. A common approach is to model the entries using an exponential-family distribution while imposing low-rank structure on the natural-parameter matrix. This leads to a broad class of generalized linear low-rank models, including logistic, Poisson and Gamma low-rank models \citep{udell2016}.

Suppose the observations $M_{ij}$ follow an exponential-family distribution and let
$X\in\mathbb{R}^{n\times p}$ denote the corresponding matrix of natural parameters. We assume that
$
\operatorname{rank}(X)\le k,
$
leading to a generalized linear low-rank model with negative log-likelihood $\ell(X)$. In general, $\ell(X)$ is not quadratic, making the resulting optimization problem difficult to solve under the low-rank constraint. To address this challenge, we adopt the iteratively reweighted least squares (IRLS) algorithm, a standard optimization technique for generalized linear models \citep{McCullagh1989}. At each iteration, the objective function is replaced by its second-order Taylor approximation around the current iterate, reducing the optimization to a WLRMA problem.

Given the current estimate $X^{(t)}$, the second-order Taylor approximation of $\ell(X)$ around $X^{(t)}$ is
\[
\ell(X)
\approx
\ell(X^{(t)})
+
\sum_{i,j}
\nabla_{ij}(X^{(t)})
(X_{ij}-X_{ij}^{(t)})
+
\frac12
\sum_{i,j}
\nabla_{ij}^2(X^{(t)})
(X_{ij}-X_{ij}^{(t)})^2,
\]
where $\nabla(X^{(t)})$ and $\nabla^2(X^{(t)})$ denote the matrices of first and second partial derivatives of $\ell(X)$ evaluated at $X^{(t)}$. Completing the square, and ignoring additive and positive multiplicative constants, yields the approximation
\[
\ell(X)
\approx
\frac 12 
\|
\sqrt{W^{(t)}}*(Z^{(t)}-X)
\|_F^2
\quad\text{with}\quad
W^{(t)}=\nabla^2(X^{(t)})
\quad\text{and}\quad
Z^{(t)}=X^{(t)}-
\frac{\nabla(X^{(t)})}
{\nabla^2(X^{(t)})},
\]
where the division is performed elementwise. Here, $W^{(t)}$ denotes the weight matrix and $Z^{(t)}$ the corresponding working response.
Consequently, each iteration reduces to solving a weighted low-rank matrix approximation problem. The resulting algorithm, which uses WLRMA as its computational building block, is summarized in Algorithm~\ref{alg:glm}.

\begin{algorithm}
\caption{GLRMA}
\label{alg:glm}
\begin{algorithmic}[1]
\Require
data matrix $M$, target rank $k$
\State \textbf{Initialize:} $X^{(0)}=0$
\For{$t=0,1,\ldots$ until convergence}

\State 
$
W^{(t)}=\nabla^2(X^{(t)})
$
\State 
$Z^{(t)}
=
X^{(t)}
-
\frac{\nabla(X^{(t)})}{\nabla^2(X^{(t)})}.
$

\State Compute $X^{(t+1)}$  by solving~\eqref{eq:wlrma} with data  $Z^{(t)}$ and weights $W^{(t)}$

\EndFor

\Ensure Natural parameter estimate $X$

\end{algorithmic}
\end{algorithm}

We illustrate this general framework by modeling the missingness pattern in the MovieLens dataset. Let
\[
M_{ij}=
\begin{cases}
1,&\text{if user }i\text{ rated movie }j,\\
0,&\text{otherwise},
\end{cases}
\]
and assume
$
M_{ij}\sim\mathrm{Bernoulli}(P_{ij}),
$
where
$
\log\left(\frac{P_{ij}}{1-P_{ij}}\right)=X_{ij}.
$
For the Bernoulli model, the weight matrix and working response become
\[
W^{(t)}
=
P^{(t)}*
\left(1-P^{(t)}\right)
\quad\text{and}\quad
Z^{(t)}
=
X^{(t)}
+
\frac{
M-P^{(t)}
}{
W^{(t)}
},
\]
where
$
P^{(t)}
=
\frac{\exp(X^{(t)})}
{1+\exp(X^{(t)})}.
$

\begin{figure}[h]
  \centering
  \begin{subfigure}{0.6\textwidth}
    \includegraphics[width=\textwidth]{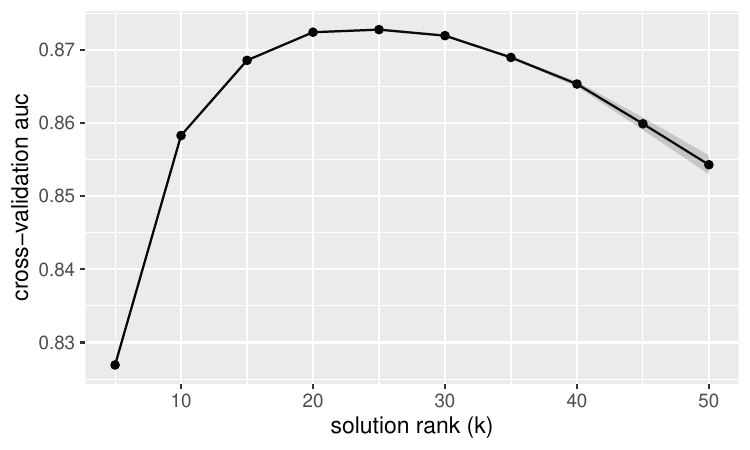}
  \end{subfigure}
  \begin{subfigure}{0.37\textwidth}
    \includegraphics[width=\textwidth]{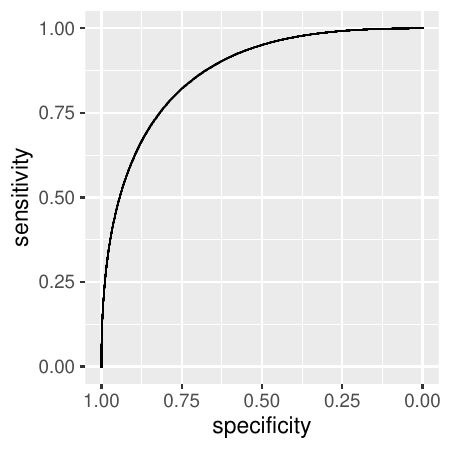}
  \end{subfigure}
  \hfill
  \begin{subfigure}{0.49\textwidth}
    \includegraphics[width=\textwidth]{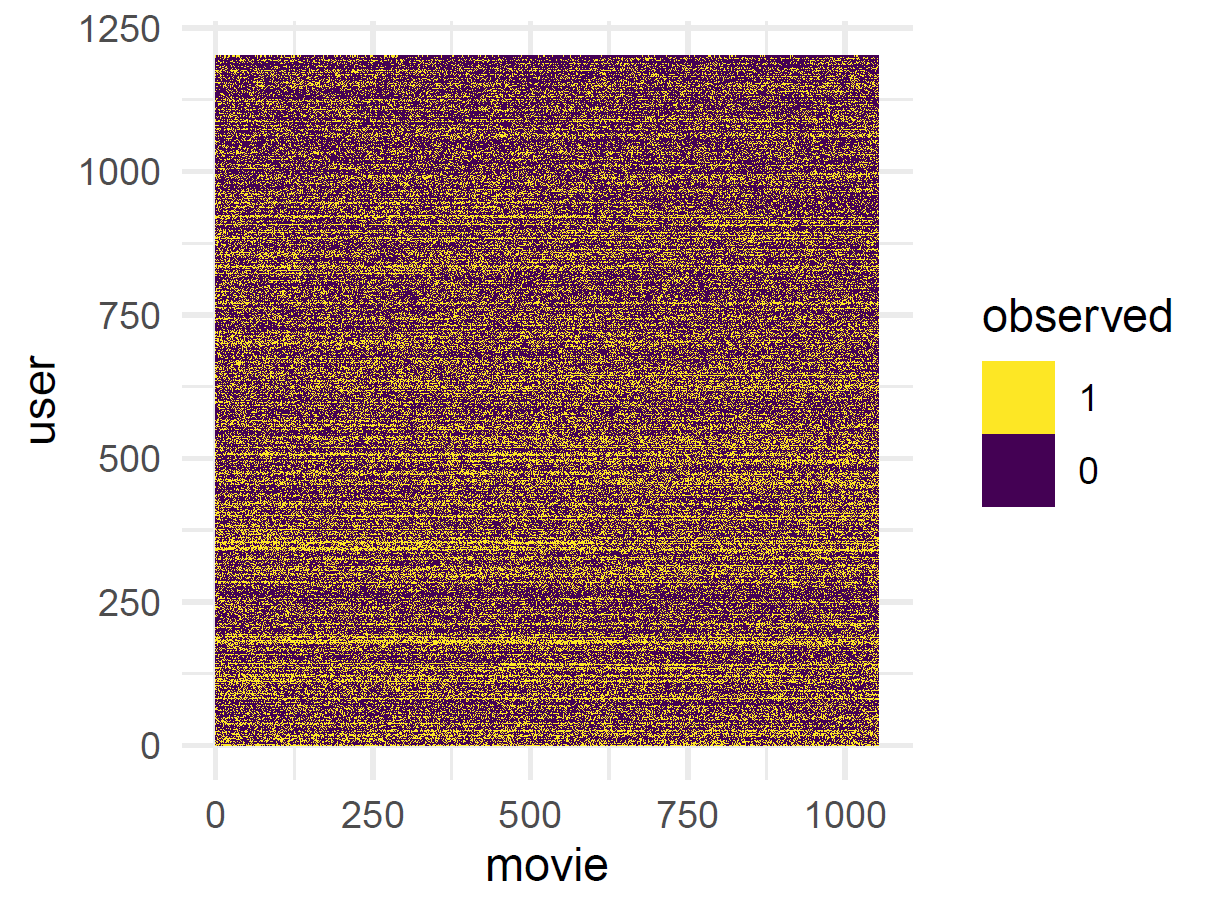}
  \end{subfigure}
  \begin{subfigure}{0.49\textwidth}
    \includegraphics[width=\textwidth]{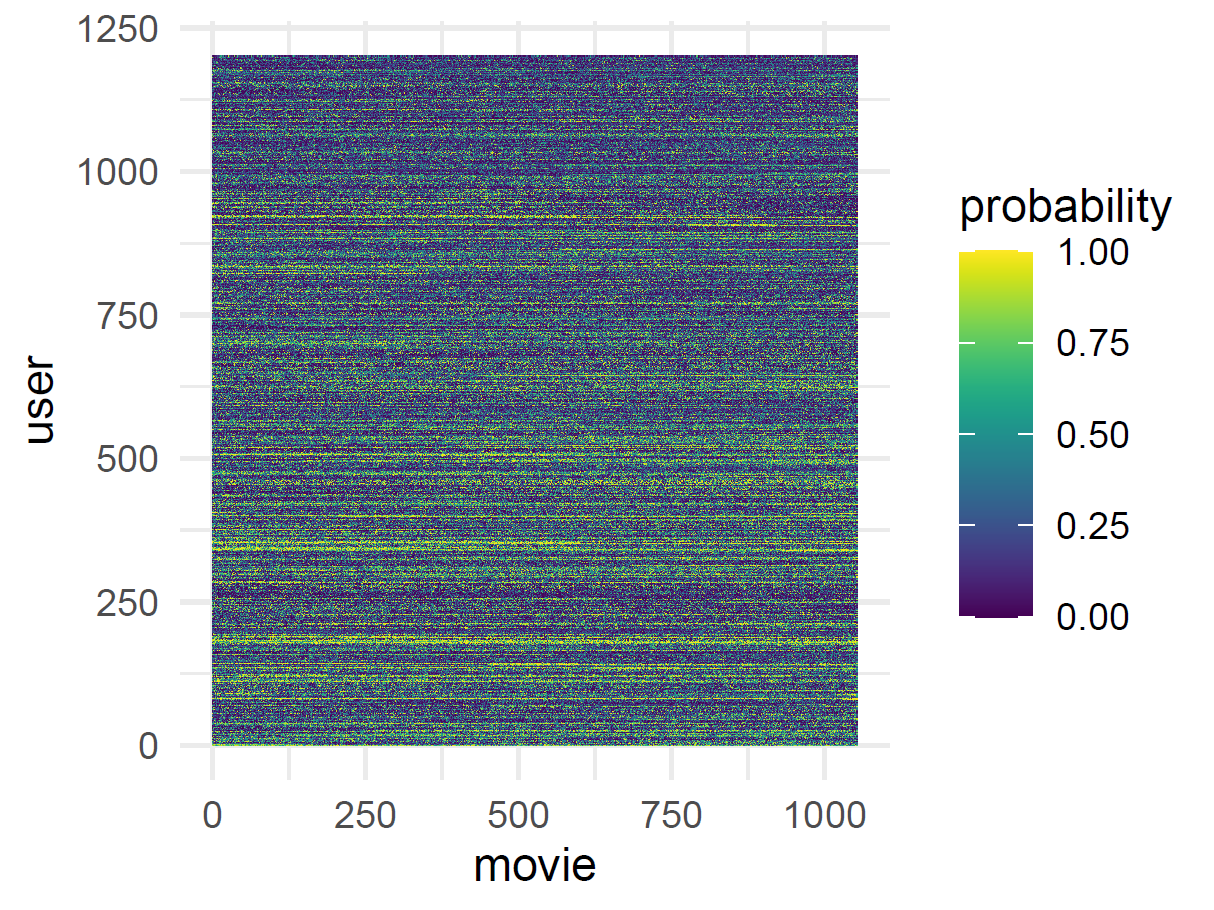}
  \end{subfigure}
 \caption{Illustration of the logistic low-rank model on the MovieLens dataset. Top-left: 10-fold cross-validation AUC versus rank. Top-right: ROC curve. Bottom-left: observed missingness pattern. Bottom-right: estimated observation probabilities.}
  \label{fig:realdata:logit}
\end{figure}

We fit the logistic low-rank model using the algorithm described above. The approximation rank is selected via 10-fold cross-validation over the grid $k=5,10,\ldots,50$, using the AUC on the held-out folds as the performance metric. The resulting cross-validation curve is displayed in the top-left panel of Figure~\ref{fig:realdata:logit}. The highest average AUC, equal to 0.873, is achieved at rank $k=25$.
Using the selected rank, we refit the model on the full dataset and obtain the estimated probability matrix $\widehat P$, shown in the bottom-right panel of Figure~\ref{fig:realdata:logit}. Comparison with the observed missingness pattern demonstrates that the fitted model accurately captures its large-scale structure despite using a relatively low-rank representation. The corresponding ROC curve, presented in the top-right panel of Figure~\ref{fig:realdata:logit}, further illustrates the strong predictive performance of the fitted model.

\section{Discussion}
\label{sec:discussion}

We revisited weighted low-rank matrix approximation through the lens of projected and proximal gradient methods. This perspective unifies the optimization of both rank-constrained and nuclear-norm formulations and naturally enables the use of acceleration techniques, including Nesterov and Anderson acceleration. The resulting algorithms substantially improve convergence speed while preserving the simplicity of the underlying optimization framework.

For high-dimensional problems, we extended the alternating least-squares framework and the sparse-plus-low-rank formulation to the weighted setting, thereby avoiding repeated singular value decompositions and making the proposed methods applicable to large-scale datasets. This factorized formulation also motivated the notion of an effective rank, which provides a more practically meaningful measure of model complexity than the algebraic rank and can be used for model selection.

Beyond optimization, we showed that weighted low-rank matrix approximation serves as a computational building block for generalized linear low-rank models. In particular, we propose an algorithm that reduces parameter estimation to a sequence of weighted low-rank approximation problems, allowing the same optimization machinery to be applied across a broad class of statistical models. The MovieLens examples illustrated these ideas for matrix completion, heteroscedastic Gaussian low-rank modeling, and logistic low-rank modeling.

Several directions for future work remain. From a theoretical perspective, it would be of interest to establish convergence guarantees for the accelerated rank-constrained algorithms and to further understand the theoretical and computational properties of the proposed ALS-based approximation for high-dimensional problems. From an applied perspective, promising extensions include distributed and online algorithms for massive datasets, adaptive or data-driven weighting schemes, and generalizations of the proposed optimization framework to weighted tensor approximation.

\section{Data and Code Availability}
The MovieLens dataset is available 
from \href{https://grouplens.org/datasets/movielens}{the GroupLens Research} website.
The proposed methods are implemented in the R package \texttt{WLRMA}; the software is available
from GitHub 
(\href{https://github.com/ElenaTuzhilina/WLRMA}{
https://github.com/ElenaTuzhilina/WLRMA}).

\section{Competing interests}
The authors declare no competing interests.

\section{Funding}
E.T. was supported by Natural Sciences and Engineering Research Council of Canada under grant RGPIN-2023-04727; the University of Toronto Data Science Institute Catalyst grant; and the University of Toronto McLaughlin Center under grant MC-2023-05.  T. H. was partially supported by grants DMS-2013736 and IIS
1837931 from the National Science Foundation, and grant 5R01 EB
001988-21 from the National Institutes of Health.

\newpage
\begin{appendices}

\section{Additional plots}
\label{app:simulation}

\subsection{Initialization}
\label{app:simu:init}

The non-convex nature of the rank-constrained WLRMA problem implies that the choice of the starting value $X^{(0)}$ may substantially affect the convergence behavior of the algorithm. We compare four initialization strategies:
\begin{enumerate}
    \item the zero matrix, $X^{(0)}=0$;
    \item a random full-rank matrix with i.i.d.\ standard normal entries;
    \item a random rank-$k$ matrix,
    $X^{(0)}=A^{(0)}{B^{(0)}}^\top,$
    where $A^{(0)}\in\mathbb{R}^{n\times k}$ and
    $B^{(0)}\in\mathbb{R}^{p\times k}$ have i.i.d.\ standard normal entries;
    \item a warm start, $X^{(0)}=\operatorname{SVD}_k(M)$.
\end{enumerate}

\begin{figure}[h!]
  \centering
  \includegraphics[width=0.9\textwidth]{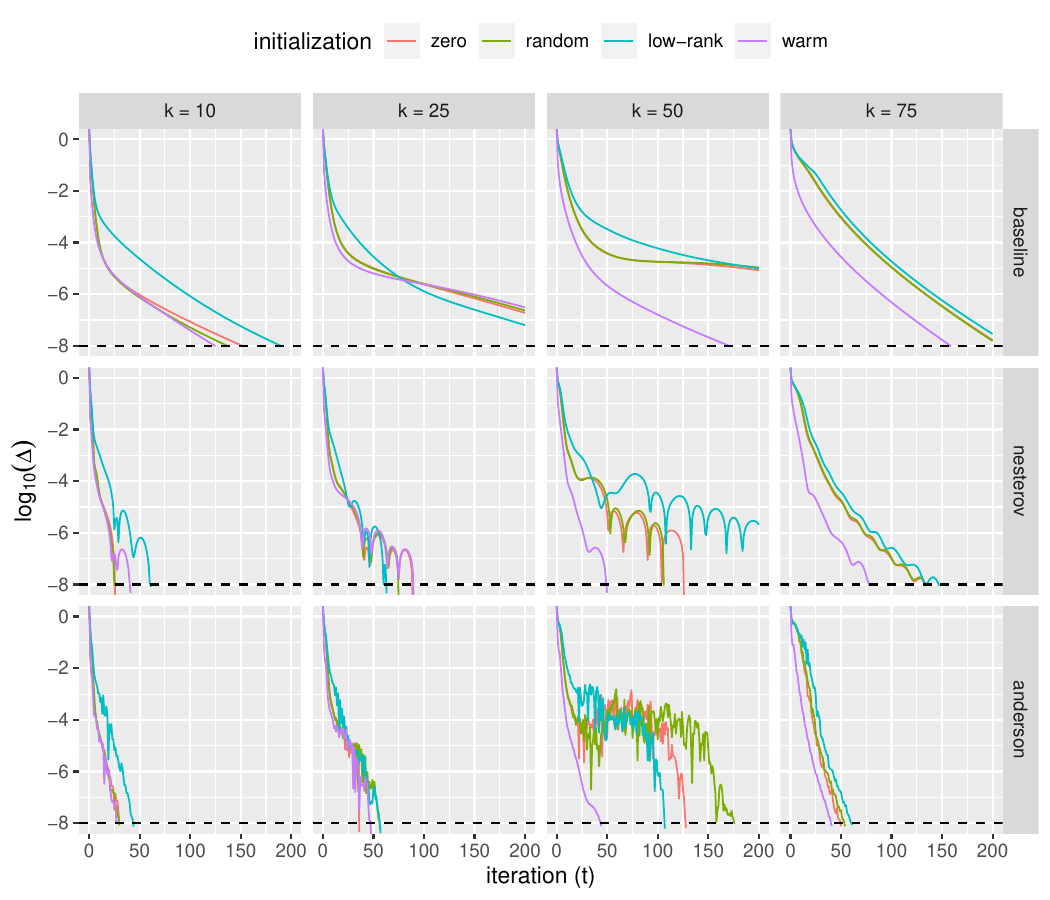}
  \caption{Effect of the initialization on the convergence of the rank-constrained WLRMA algorithm. The curves show the relative objective change $\Delta^{(t)}$ versus iteration number for different initialization strategies. Columns correspond to different solution ranks $k$, while rows correspond to the baseline, Nesterov, and Anderson algorithms.}
  \label{fig:simulation:init}
\end{figure}

As illustrated in Figure~\ref{fig:simulation:init}, the warm-start initialization consistently leads to faster convergence and substantially improves the stability of the accelerated methods.

\clearpage

\subsection{Rank}
\label{app:simu:rank}

\begin{figure}[h!]
  \centering
  \includegraphics[width=\textwidth]{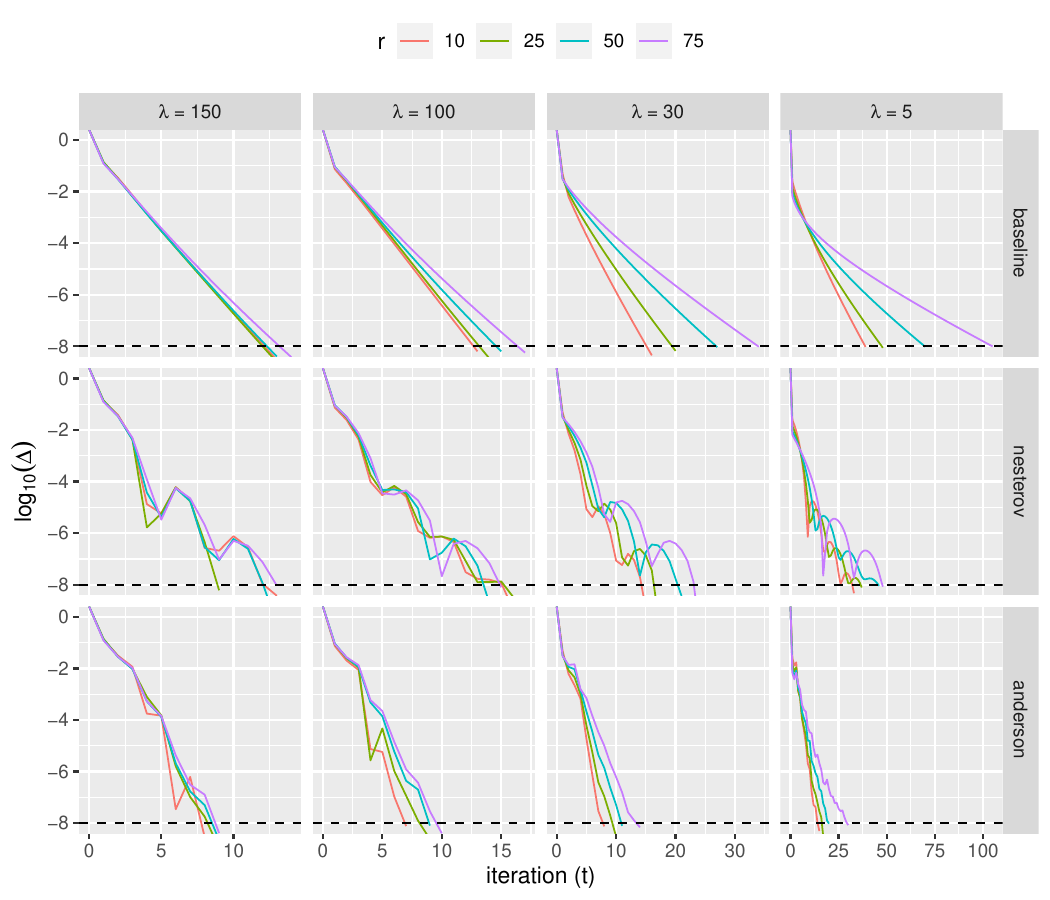}
  \caption{Effect of the true rank of the simulated matrix on the convergence of the convex WLRMA algorithm. The curves show the relative objective change $\Delta^{(t)}$ versus iteration number for different ranks. Columns correspond to different values of the penalty parameter $\lambda$, while rows correspond to the baseline, Nesterov, and Anderson algorithms.}
  \label{fig:simulation:rank}
\end{figure}

We investigate the effect of the true rank of the simulated matrix on the convergence of the convex WLRMA algorithm. As illustrated in Figure~\ref{fig:simulation:rank}, lower-rank matrices generally require fewer iterations to converge.

\clearpage

\subsection{Noise level}
\label{app:simu:noise}

\begin{figure}[h!]
  \centering
  \includegraphics[width=\textwidth]{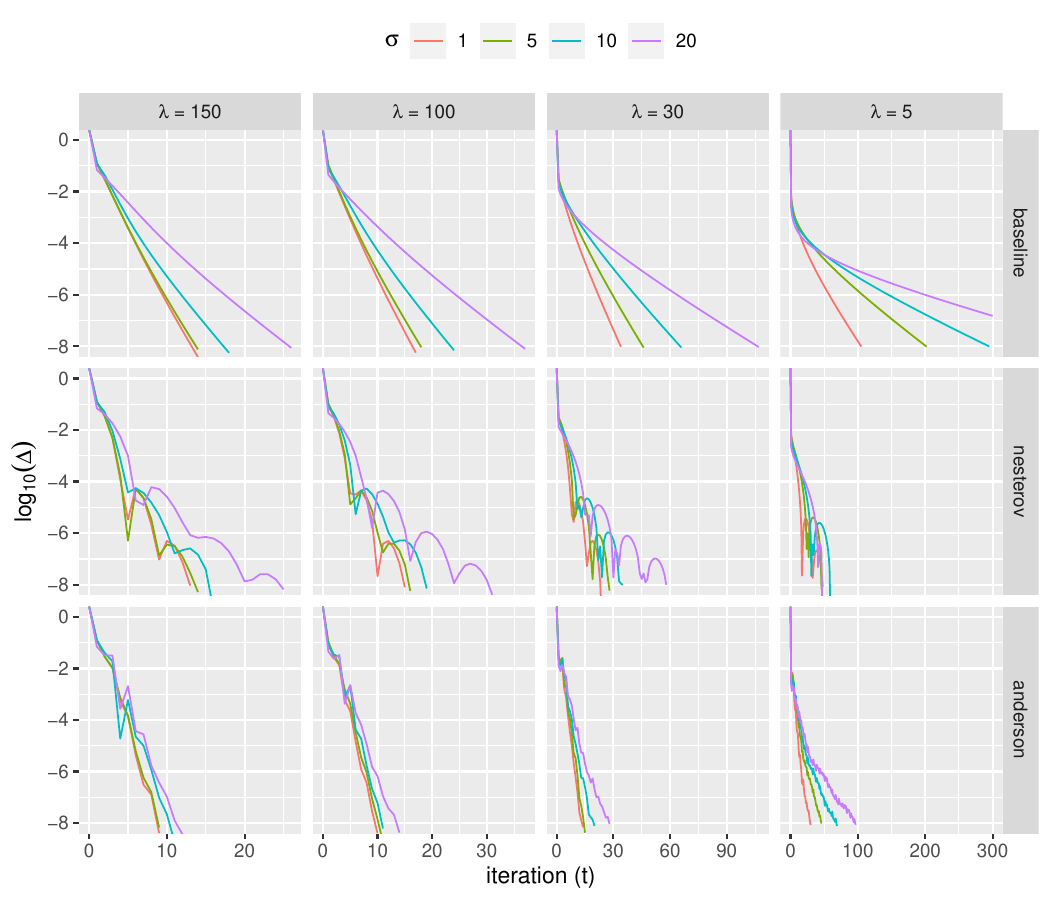}
  \caption{Effect of the noise level on the convergence of the convex WLRMA algorithm. The curves show the relative objective change $\Delta^{(t)}$ versus iteration number for different values of $\sigma$. Columns correspond to different values of the penalty parameter $\lambda$, while rows correspond to the baseline, Nesterov, and Anderson algorithms.}
  \label{fig:simulation:noise}
\end{figure}

We investigate the effect of the noise level in the simulated data on the convergence of the convex WLRMA algorithm. As illustrated in Figure~\ref{fig:simulation:noise}, lower noise levels generally require fewer iterations to converge. This behavior is expected, since a higher signal-to-noise ratio yields a more accurate low-rank approximation, making the optimization problem easier to solve.

\clearpage 

\subsection{Effective rank}
\label{app:er}

As discussed in Section \ref{sec:er}, there are two ways to define effective rank for a soft-impute solution.
When fixing $A^*$ the effective rank formula is
\begin{align}
\operatorname{er}(\lambda)
=
\frac{1}{p}
\sum_{j=1}^{p}
\operatorname{tr}\!\left[
\left(
{A^*}^{\top}D_jA^*+\lambda I
\right)^{-1}
{A^*}^{\top}D_jA^*
\right],
\qquad
D_j=\operatorname{diag}(W_{\cdot j}).
\end{align}
In contrast, one can fix $B^*$, which leads to the alternative formula
\begin{align}
\operatorname{er}(\lambda)
=
\frac{1}{n}
\sum_{i=1}^{n}
\operatorname{tr}\!\left[
\left({B^*}^{\top}D_iB^*+\lambda I\right)^{-1}
{B^*}^{\top}D_iB^*
\right],
\qquad
D_i=\operatorname{diag}(W_{i\cdot}).\end{align}
Below we compare the two proposed formulas of the effective rank via the simulation and real data experiments. As we observe from both experiments, fixing $A^*$ and $B^*$ leads to compatible results of effective rank.

\begin{figure}[h!]
  \centering
    \includegraphics[width = 0.6\textwidth]{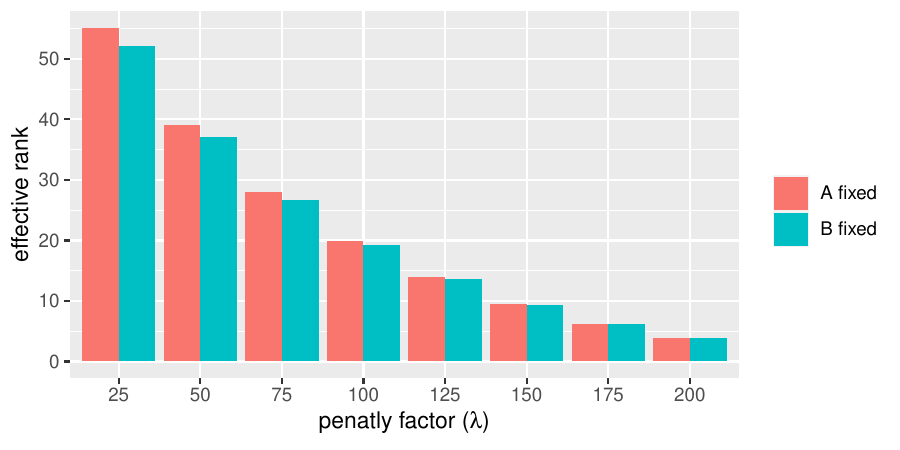}
    \caption{Comparison of two definitions of the effective rank in the simulation example. Red: $A$ fixed. Blue: $B$ fixed. Both definitions yield similar effective ranks.}
\end{figure}

\begin{figure}[h!]
  \centering
    \includegraphics[width = 0.6\textwidth]{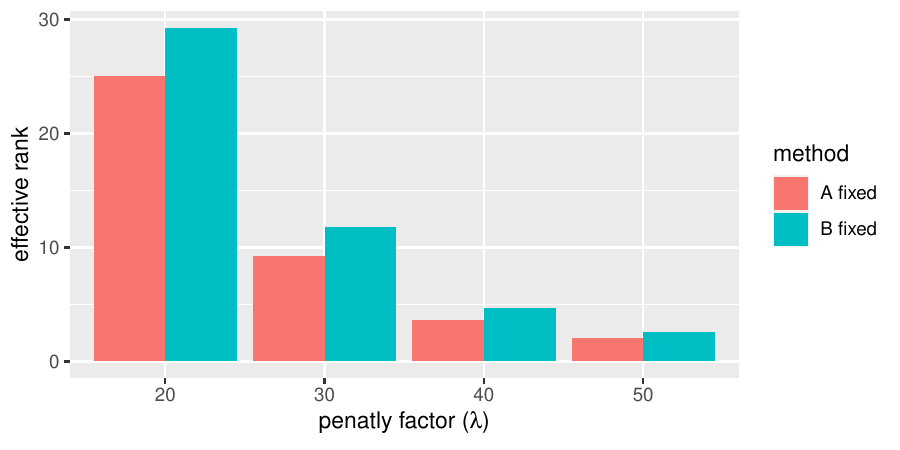}
     \caption{Comparison of two definitions of the effective rank in the MovieLens example. Red: $A$ fixed. Blue: $B$ fixed. Both definitions yield similar effective ranks.}
\end{figure}




\end{appendices}


\bibliography{refs}

\end{document}